%% file: main.tex
\documentclass{article}
\input{Prefix}

\title{\sysname:  Speculative Decoding for Batch Inference of LLM Agents}

\author{
    Xin Wang\textsuperscript{1}\thanks{Work was done during internship at Microsoft Research.}, Ziming Miao\textsuperscript{2}, Yi Zhu\textsuperscript{2}, Hui Shen\textsuperscript{3}, Zhongwei Wan\textsuperscript{1}, Fan Yang\textsuperscript{2}, Mi Zhang\textsuperscript{1} \\
    \textsuperscript{1}The Ohio State University \quad
    \textsuperscript{2}Microsoft Research \quad \textsuperscript{3}University of Michigan
}
\begin{document}

\maketitle
\begin{abstract}
\input{Sections/0_abstract}
\end{abstract}

\input{Sections/1_Introduction}

\input{Sections/2_Related_Works}

\input{Sections/3_Analysis}

\input{Sections/4_Method}

\input{Sections/5_Experiment}

\input{Sections/6_Conclusion}


\setcitestyle{numbers}
\bibliography{Reference}

\appendix
\input{Sections/Apprendix}
\end{document}

%% file: Prefix.tex
\usepackage{microtype}
\usepackage{graphicx}
\usepackage{subfigure}

\usepackage{caption}
\usepackage{xspace}
\usepackage{pifont}
\usepackage{booktabs} 
\usepackage{hyperref}
\usepackage{xcolor}
\usepackage{listings}
\usepackage{times}
\usepackage{latexsym}

\usepackage{algorithm}
\usepackage{algorithmic}

\usepackage[T1]{fontenc}

\usepackage[utf8]{inputenc}

\usepackage[final]{acl}

\usepackage{amsmath}
\usepackage{amssymb}
\usepackage{mathtools}
\usepackage{amsthm}
\usepackage{multirow}
\usepackage{graphicx}
\usepackage{lscape}
\usepackage[capitalize,noabbrev]{cleveref}
\theoremstyle{plain}

\theoremstyle{definition}

\theoremstyle{remark}

\newcommand{\sysname}{\texttt{AgentSpec}\xspace}

\definecolor{mygreen}{HTML}{009901}
\newcommand{\spup}[1]{\textcolor{mygreen}{#1}}
\newcommand{\spdown}[1]{\textcolor{red}{#1}}
\newcommand{\spone}[1]{\textcolor{gray}{#1}}
\usepackage[textsize=tiny]{todonotes}
\input{Table}

%% file: Table.tex
\newcommand{\eeTable}{{
\begin{table*}[t]
\centering
\caption{Goodput (tokens/sec) of \sysname and baselines on four different agentic workloads with three different models. The best performance is marked in bold. The speedup is measured by comparing the goodput with the normal autoregressive decoding. The relative performance gain compared to the best-performing baseline is marked in green inside bracket.}
\label{tab:e2e_comparison}
\resizebox{0.95\linewidth}{!}{
\begin{tabular}{c|c|cc|cc|cc|cc}
\toprule
\textsc{Model} & \textsc{Method}
& \multicolumn{2}{c|}{Deep Research Agent}
& \multicolumn{2}{c|}{Code Generation Agent}
& \multicolumn{2}{c|}{GAIA}
& \multicolumn{2}{c}{SWE-Bench}\\
\cmidrule(lr){3-4}\cmidrule(lr){5-6}\cmidrule(lr){7-8}\cmidrule(lr){9-10}
& & Tokens/s & Speedup & Tokens/s & Speedup & Tokens/s & Speedup & Tokens/s & Speedup\\
\midrule

\multirow{5}{*}{\rotatebox[origin=c]{90}{\textbf{Qwen-3-8B}}} & Normal
& 617.41 & \spone{$\times$ 1.00}
& 636.09 & \spone{$\times$ 1.00}
& 501.27 & \spone{$\times$ 1.00}
& 652.44 & \spone{$\times$ 1.00} \\ \cmidrule{2-10}
         & EAGLE-3
& 605.14 & \spdown{$\times$ 0.98}
& 539.01 & \spdown{$\times$ 0.85}
& 468.22 & \spdown{$\times$ 0.93}
& 523.12 & \spdown{$\times$ 0.80} \\
         & NGram
& 509.62 & \spdown{$\times$ 0.83}
& 422.58 & \spdown{$\times$ 0.66}
& 308.27 & \spdown{$\times$ 0.61}
& 408.28 & \spdown{$\times$ 0.63} \\
         & SuffixDecoding
& 598.74 & \spdown{$\times$ 0.97}
& 579.21 & \spdown{$\times$ 0.91}
& 498.16 & \spdown{$\times$ 0.99}
& 556.64 & \spdown{$\times$ 0.85} \\ \cmidrule{2-10}
         & \sysname
& \textbf{661.43} \spup{($\uparrow 9\%$)}  & \spup{\textbf{$\times$ 1.07}}
& \textbf{828.10} \spup{($\uparrow 43\%$)} & \spup{\textbf{$\times$ 1.30}}
& \textbf{537.78} \spup{($\uparrow 8\%$)}  & \spup{\textbf{$\times$ 1.07}}
& \textbf{854.69} \spup{($\uparrow 42\%$)} & \spup{\textbf{$\times$ 1.31}} \\
\midrule

\multirow{5}{*}{\rotatebox[origin=c]{90}{\textbf{GPT-OSS-20B}}} & Normal
& 201.43 & \spone{$\times$ 1.00}
& 289.26 & \spone{$\times$ 1.00}
& 247.67 & \spone{$\times$ 1.00}
& 351.22 & \spone{$\times$ 1.00} \\ \cmidrule{2-10}
            & EAGLE-3
& 125.71 & \spdown{$\times$ 0.62}
& 161.34 & \spdown{$\times$ 0.56}
& 112.16 & \spdown{$\times$ 0.45}
& 198.81 & \spdown{$\times$ 0.57} \\
            & NGram
& 158.91 & \spdown{$\times$ 0.79}
& 175.19 & \spdown{$\times$ 0.61}
& 109.22 & \spdown{$\times$ 0.44}
& 253.36 & \spdown{$\times$ 0.72} \\
            & SuffixDecoding
& 196.54 & \spdown{$\times$ 0.98}
& 295.85 & \spup{$\times$ 1.02}
& 187.79 & \spdown{$\times$ 0.76}
& 319.73 & \spdown{$\times$ 0.91} \\ \cmidrule{2-10}
            & \sysname
& \textbf{297.54} \spup{($\uparrow 51\%$)} & \spup{\textbf{$\times$ 1.48}}
& \textbf{584.43} \spup{($\uparrow 104\%$)} & \spup{\textbf{$\times$ 2.02}}
& \textbf{295.76} \spup{($\uparrow 57\%$)} & \spup{\textbf{$\times$ 1.19}}
& \textbf{595.20} \spup{($\uparrow 86\%$)} & \spup{\textbf{$\times$ 1.69}} \\
\midrule

\multirow{5}{*}{\rotatebox[origin=c]{90}{\textbf{DSK-Distill-8B}}} & Normal
& 987.66  & \spone{$\times$ 1.00}
& 1230.28 & \spone{$\times$ 1.00}
& 823.39  & \spone{$\times$ 1.00}
& 1432.27 & \spone{$\times$ 1.00} \\ \cmidrule{2-10}
       & EAGLE-3
& 965.43  & \spdown{$\times$ 0.98}
& 1119.29 & \spdown{$\times$ 0.91}
& 864.43  & \spup{$\times$ 1.05}
& 1098.82 & \spdown{$\times$ 0.77} \\
       & NGram
& 752.66  & \spdown{$\times$ 0.76}
& 797.18  & \spdown{$\times$ 0.65}
& 725.29  & \spdown{$\times$ 0.88}
& 1014.47 & \spdown{$\times$ 0.71} \\
       & SuffixDecoding
& 995.14  & \spup{$\times$ 1.01}
& 1041.29 & \spdown{$\times$ 0.85}
& 820.09  & \spone{$\times$ 1.00}
& 1498.69 & \spup{$\times$ 1.05} \\ \cmidrule{2-10}
       & \sysname
& \textbf{1294.65} \spup{($\uparrow 30\%$)} & \spup{\textbf{$\times$ 1.31}}
& \textbf{1671.34} \spup{($\uparrow 49\%$)} & \spup{\textbf{$\times$ 1.36}}
& \textbf{924.47}  \spup{($\uparrow 7\%$)}  & \spup{\textbf{$\times$ 1.12}}
& \textbf{2281.87} \spup{($\uparrow 52\%$)} & \spup{\textbf{$\times$ 1.59}}\\
\bottomrule
\end{tabular}
}
\end{table*}
}}

\newcommand{\mimoTable}{{
\begin{table*}[ht]
\centering
\caption{Goodput (tokens/sec) of \sysname and baselines, including MTP, on MiMo-7B. The best performance is marked in bold. The speedup is measured by comparing the goodput with the normal autoregressive decoding. The relative performance gain compared to the best-performing baseline is marked in green inside bracket.}
\label{tab:mimo_usaco}
\resizebox{1\linewidth}{!}{
\begin{tabular}{c| c| cc cc cc cc cc}
\toprule
\textsc{Model} & \textsc{Method}
& \multicolumn{2}{c}{Code Gen.(Bronze)}
& \multicolumn{2}{c}{Code Gen.(Silver)}
& \multicolumn{2}{c}{Code Gen.(Gold)}
& \multicolumn{2}{c}{Code Gen.(Platinum)}
& \multicolumn{2}{c}{Code Gen.(All)} \\
\cmidrule(lr){3-4}\cmidrule(lr){5-6}\cmidrule(lr){7-8}\cmidrule(lr){9-10}\cmidrule(lr){11-12}
& & Tokens/s & Speedup & Tokens/s & Speedup & Tokens/s & Speedup & Tokens/s & Speedup & Tokens/s & Speedup \\
\midrule

\multirow{5}{*}{\rotatebox[origin=c]{90}{\textbf{MiMo-7B}}} & Normal
& 2768.67 & \spone{$\times$ 1.00}
& 3093.50 & \spone{$\times$ 1.00}
& 2532.30 & \spone{$\times$ 1.00}
& 1141.64 & \spone{$\times$ 1.00}
& 2572.16 & \spone{$\times$ 1.00} \\ \cmidrule{2-12}
        & NGram
& 1322.92 & \spdown{$\times$ 0.48}
& 1399.76 & \spdown{$\times$ 0.45}
& 1383.92 & \spdown{$\times$ 0.55}
& 990.87   & \spdown{$\times$ 0.87}
& 1306.92 & \spdown{$\times$ 0.51} \\
        & SuffixDecoding
& 1609.41 & \spdown{$\times$ 0.58}
& 1624.98 & \spdown{$\times$ 0.53}
& 1599.63 & \spdown{$\times$ 0.63}
& 1235.12 & \spup{$\times$ 1.08}
& 1653.26 & \spdown{$\times$ 0.64} \\
        & MTP
& 1924.46 & \spdown{$\times$ 0.70}
& 2001.28 & \spdown{$\times$ 0.65}
& 2034.30 & \spdown{$\times$ 0.80}
& 1349.10 & \spup{$\times$ 1.18}
& 1875.18 & \spdown{$\times$ 0.73} \\ \cmidrule{2-12}
        & \sysname
& \textbf{3324.71} \spup{($\uparrow 73\%$)} & \spup{$\times$ \textbf{1.20}}
& \textbf{3751.44} \spup{($\uparrow 87\%$)} & \spup{$\times$ \textbf{1.21}}
& \textbf{2925.40} \spup{($\uparrow 44\%$)} & \spup{$\times$ \textbf{1.16}}
& \textbf{1527.14} \spup{($\uparrow 13\%$)} & \spup{$\times$ \textbf{1.34}}
& \textbf{3318.99} \spup{($\uparrow 77\%$)} & \spup{$\times$ \textbf{1.29}} \\

\bottomrule
\end{tabular}
}
\end{table*}
}}

\newcommand{\specbenchTable}{{
\begin{table*}[ht]
\centering
\caption{Goodput (tokens/sec) of \sysname and baselines on Spec-Bench on Qwen-3-8B. The best performance is marked in bold. The speedup is measured by comparing the goodput with the normal autoregressive decoding.}
\label{tab:qwen_specbench}
\resizebox{1\linewidth}{!}{
\begin{tabular}{c| c| cc cc cc cc cc cc cc}
\toprule
\textsc{Model} & \textsc{Method}
& \multicolumn{2}{c}{MT Conversation}
& \multicolumn{2}{c}{Question Answering}
& \multicolumn{2}{c}{Summarization}
& \multicolumn{2}{c}{Translation}
& \multicolumn{2}{c}{Math Reasoning}
& \multicolumn{2}{c}{Retrieval Augmented}
& \multicolumn{2}{c}{All}\\
\cmidrule(lr){3-4}\cmidrule(lr){5-6}\cmidrule(lr){7-8}\cmidrule(lr){9-10}
\cmidrule(lr){11-12}\cmidrule(lr){13-14}\cmidrule(lr){15-16}
& & Tokens/s & Speedup & Tokens/s & Speedup & Tokens/s & Speedup
& Tokens/s & Speedup & Tokens/s & Speedup & Tokens/s & Speedup & Tokens/s & Speedup \\
\midrule

\multirow{5}{*}{\rotatebox[origin=c]{90}{\textbf{Qwen-3-8B}}}
&           Normal
& 939.50 & \spone{$\times$1.00}
& 860.97 & \spone{$\times$1.00}
& 1277.50 & \spone{$\times$1.00}
& 2013.02 & \spone{$\times$1.00}
& 1064.78 & \spone{$\times$1.00}
& 191.60 & \spone{$\times$1.00}
& 697.42 & \spone{$\times$1.00} \\\cmidrule{2-16}
&           EAGLE-3
& 1005.63 & \spup{$\times$1.07}
& 895.25  & \spup{$\times$1.04}
& 1552.62 & \spup{$\times$1.22}
& 2319.01 & \spup{$\times$1.15}
& 1045.28 & \spdown{$\times$0.98}
& 168.21  & \spdown{$\times$0.88}
& 783.25  & \spup{$\times$1.12} \\
&           NGram
& 581.26 & \spdown{$\times$0.62}
& 351.28 & \spdown{$\times$0.41}
& 1082.57 & \spdown{$\times$0.85}
& 2166.37 & \spup{$\times$1.08}
& 959.38 & \spdown{$\times$0.90}
& 263.11 & \spup{$\times$1.37}
& 545.39 & \spdown{$\times$0.78} \\
&           SuffixDecoding
& 1007.08 & \spup{$\times$1.07}
& 873.12  & \spup{$\times$1.01}
& 1213.63 & \spdown{$\times$0.95}
& 2141.79 & \spup{$\times$1.06}
& 1121.48 & \spup{$\times$1.05}
& 212.19 & \spup{$\times$1.10}
& 712.79 & \spup{$\times$1.02} \\\cmidrule{2-16}
&           \sysname
& \textbf{1128.09} & \spup{$\times$ \textbf{1.20}}
& \textbf{929.85} & \spup{$\times$ \textbf{1.08}}
& \textbf{1481.90} & \spup{$\times$ \textbf{1.16}}
& \textbf{2294.12} & \spup{$\times$ \textbf{1.14}}
& \textbf{1075.43} & \spup{$\times$ \textbf{1.01}}
& \textbf{267.14} & \spup{$\times$ \textbf{1.40}}
& \textbf{796.01} & \spup{$\times$ \textbf{1.14}} \\
\bottomrule
\end{tabular}
}
\end{table*}
}}

\newcommand{\sensitivityTable}{{
\begin{table}[t]
\centering
\caption{Goodput of variants of \sysname and baselines on Deep Research Agent and Code Generation Agent workloads. The best performance is marked in bold. The speedup is measured by comparing the goodput with the normal autoregressive decoding.}
\label{tab:sensitivity}
\resizebox{\linewidth}{!}{
\begin{tabular}{c|c|cc|cc}
\toprule
\textsc{Model} & \textsc{Method}
& \multicolumn{2}{c|}{Deep Research Agent}
& \multicolumn{2}{c}{Code Generation Agent} \\
\cmidrule(lr){3-4}\cmidrule(lr){5-6}
& & Tokens/s & Speedup & Tokens/s & Speedup \\
\midrule
\multirow{7}{*}{\rotatebox[origin=c]{90}{\textsc{\textbf{Qwen-3-8B}}}}
& Normal         & 617.41 & \spone{$\times$ 1.00}                 & 636.09 & \spone{$\times$ 1.00} \\ \cmidrule(lr){2-6}
& EAGLE-3         & 605.14 & \spdown{$\times$ 0.98} & 539.01 & \spdown{$\times$ 0.85} \\
& NGram          & 509.62 & \spdown{$\times$ 0.83} & 422.58 & \spdown{$\times$ 0.66} \\
& SuffixDecoding & 598.74 & \spdown{$\times$ 0.97} & 579.21 & \spdown{$\times$ 0.91} \\
\cmidrule(lr){2-6}
& \textsc{\sysname\texttt{(S)}}
& \textbf{647.12}~\spup{($\uparrow 9\%$)}
& \spup{$\times$ \textbf{1.05}}
& \textbf{795.54}~\spup{($\uparrow 37\%$)}
& \spup{$\times$ \textbf{1.25}} \\
& \textsc{\sysname\texttt{(R)}}
& \textbf{623.33}~\spup{($\uparrow 3\%$)}
& \spup{$\times$ \textbf{1.01}}
& \textbf{702.28}~\spup{($\uparrow 21\%$)}
& \spup{$\times$ \textbf{1.10}} \\  \cmidrule{2-6}
& \textsc{\sysname\texttt{(N)}}
& \textbf{661.08}~\spup{($\uparrow 10\%$)}
& \spup{$\times$ \textbf{1.07}}
& \textbf{809.12}~\spup{($\uparrow 40\%$)}
& \spup{$\times$ \textbf{1.27}} \\   \cmidrule{2-6}
& \textsc{\sysname}
& \textbf{661.43}~\spup{($\uparrow 10\%$)}
& \spup{$\times$ \textbf{1.07}}
& \textbf{828.10}~\spup{($\uparrow 43\%$)}
& \spup{$\times$ \textbf{1.30}} \\
\bottomrule
\end{tabular}
}
\end{table}
}}

\newcommand{\rejTable}{{
\begin{table}[t]
\centering
\caption{Average per-request rejection rate \spup{($\downarrow$)}, defined as number of rejected tokens divided by number of draft tokens, for different speculative decoding algorithms on Reflexion agentic workflow.}
\label{tab:rejection_rate}
\resizebox{1\linewidth}{!}{
\begin{tabular}{c|ccc|c}
\toprule
\textsc{Metric} & EAGLE-3 & NGram & SuffixDecoding  & \sysname\\
\midrule
Rejection Rate & 53.3\% & 88.7\% & 65.9\%  & 26.4\% \spup{($\downarrow 50\%$)}\\
\bottomrule
\end{tabular}
}
\end{table}
}}

\newcommand{\breakdownTable}{{
\begin{table}[t]
\centering
\caption{Execution breakdown of \sysname and baselines on Code Gen agentic Workload. The best performance is marked in bold. }
\label{tab:runtime_breakdown}
\resizebox{\linewidth}{!}{%
\begin{tabular}{l|cccc}
\toprule
\textsc{Time (minutes)} & EAGLE-3 & NGram & SuffixDecoding & \sysname \\
\midrule
Draft        & 5.72  & \textbf{0.68}  & 1.41  & 1.74  \\\midrule
Verification & 34.24 & 39.82 & 33.12 & \textbf{26.11}~\spup{($\downarrow 21\%$)} \\\midrule
Overall      & 39.96 & 40.50 & 34.53 & \textbf{27.85}~\spup{($\downarrow 19\%$)} \\
\bottomrule
\end{tabular}%
}
\end{table}
}}

\newcommand{\GPUTable}{{
\centering
\caption{Speedup of \sysname on different GPU platforms on GPT-OSS-20B across agent workloads.}
\label{tab:hardware_speedup}
\resizebox{\linewidth}{!}{%
\begin{tabular}{l|cccc}
\toprule
\textbf{Speedup} & \textbf{CodeGen} & \textbf{DeepResearch} & \textbf{SWE-Bench} & \textbf{GAIA} \\
\midrule
1$\times$A100 80G & 2.02$\times$ & 1.48$\times$ & 1.69$\times$ & 1.19$\times$ \\
1$\times$H100 80G & 2.28$\times$ & 1.67$\times$ & 1.92$\times$ & 1.54$\times$ \\
\bottomrule
\end{tabular}%
}
}}

%% file: Sections/0_Abstract.tex
Large language model (LLM)–based agent applications often incur high response time. Speculative decoding is a promising solution to improve the inference efficiency of LLM agents without impacting generation quality. 
However, existing speculative decoding algorithms exhibit substantial speed degradation as batch sizes grow, limiting their practicality to deploy in real-world agent applications.
In this work, we first present a systematic analysis of speculative decoding for LLM agents and identify two dominant factors of speedup degradation: high rejection rate of speculative tokens, and under-utilization of dynamic token budgets.
Motivated by these findings, we propose \sysname, a speculative decoding algorithm that addresses the limitations of existing methods for LLM agents. \sysname incorporates structure-isolated drafting that constrains speculation to semantically coherent segments of the agent workflow, reducing the drafts of irrelevant semantic paths and achieving an extremely low rejection rate. Moreover, \sysname adopts redundancy-aware budget allocation that exploits agent-level information to better utilize the dynamically-free token budget during the agent inference.
We implement and evaluate \sysname on five different workloads and four different models from four different LLM families in vLLM. Our results demonstrate the superiority of \sysname over state-of-the-art.
%

%% file: Sections/1_Introduction.tex
\section{Introduction}
\label{sec:introduction}
Large language model (LLM)–based agent applications have emerged as a powerful paradigm for solving complex tasks that require multi-step reasoning, tool invocation, and environment interaction~\citep{DBLP:conf/coling/Li25,DBLP:journals/corr/abs-2503-21460}. However, in practical deployments, serving LLM agents often incurs high inference cost due to its long and multi-round generation~\citep{DBLP:journals/tmlr/Wan0LA0LQYZZC024,wang2024iot,DBLP:journals/corr/abs-2502-13965, DBLP:journals/corr/abs-2508-02694}, motivating the need for efficient methods for LLM agents.

Speculative decoding is a promising technique to reduce response time for LLM agents without impacting the generation quality~\citep{DBLP:conf/acl/XiaYDW00L0S24}. Previous works have shown notable speedups under small batch sizes~\citep{DBLP:journals/corr/abs-2503-01840, saxena2023prompt, oliaro2025suffixdecoding, DBLP:journals/corr/abs-2305-09781, 10.1145/3772052.3772239}. However, modern LLM serving systems~\citep{DBLP:conf/sosp/KwonLZ0ZY0ZS23, DBLP:conf/nips/ZhengYXS0YCKSGB24} typically operate with large batch sizes to maximize hardware utilization, where state-of-the-art speculative decoding algorithms suffer substantial speed degradation, limiting their effectiveness for real-world agent applications.

To further understand such limitation, we conduct a systematic analysis of speculative decoding for LLM agents and conclude two dominant efficiency bottlenecks: 
\ding{182} \textbf{High rejection rate of speculative tokens:} Existing speculative decoding algorithms~\citep{DBLP:journals/corr/abs-2503-01840, saxena2023prompt, oliaro2025suffixdecoding} incur a high rejection rate when applied to LLM agent workloads, resulting in substantial verification overhead for rejected tokens. As batch size increases, such overhead grows rapidly and can quickly outweigh the time saved by accepted tokens, leading to severe speedup degradation.
\ding{183} \textbf{Under-utilization of dynamic token budgets:} The amount of token budget that can be used for speculation varies dynamically across requests and batches. However, existing approaches allocate speculative budgets either uniformly~\citep{DBLP:journals/corr/abs-2305-09781, DBLP:conf/acl/WuZVPRL25, DBLP:conf/iclr/SadhukhanCCTLSY25} or in a coarse-grained manner~\citep{10.1145/3772052.3772239}, leading to inefficient use of available token budgets and limited speedup for agentic workloads.

In this paper, we propose \sysname, a new speculative decoding algorithm that addresses the limitations of existing methods for LLM agents from prior approaches in two key aspects.
\ding{182} \textbf{Structure-isolated drafting:} \sysname constrains speculation to semantically coherent segments of the agent workflow. By avoiding speculation across heterogeneous execution stages, \sysname reduces the generation of irrelevant speculative paths and achieves a substantially lower rejection rate.
\ding{183} \textbf{Redundancy-aware budget allocation:} \sysname introduces a redundancy metric to guide the allocation of dynamically available token budgets across requests, improving overall inference efficiency for LLM agents under various batch sizes.

We implement \sysname in vLLM and compare it with four speculative decoding methods, including NGram, EAGLE-3, SuffixDecoding as well as state-of-the-art method MTP. To demonstrate the generability of \sysname, we conduct our evaluation on $4$ models from four different LLM families (Qwen, DeepSeek, GPT-OSS, and MiMo) and on $4$ different agentic workloads, including $2$ workflow-based (Code Generation and Deep Research) and $2$ model-based (SWE-Bench and GAIA).
We highlight five of our findings:
(1) \sysname consistently outperforms EAGLE-3, NGram and SuffixDecoding across all 4 agentic benchmarks and 4 different LLM families. 
%
(2) All existing speculative decoding algorithms suffer from severe speedup degradation for batch inference on agent workload and even become slower than normal autoregressive decoding. However, \sysname consistently achieves faster generation than autoregressive decoding, with at most 2.02 $\times$ speedup.
(3) \sysname is also able to accelerate the batch inference on non-agentic workloads. Specifically, \sysname achieves 1.14 $\times$ speedup on Spec-Bench dataset with at most 1.40 $\times$ speed acceleration on its subset.
(4) Moreover, \sysname even achieves a higher speedup compared with Multi-Token Prediction (MTP) on MiMo-7B.
(5) Lastly, \sysname ensures a robust efficiency under various maximum batch size and even achieves a better speedup with none thinking mode.

%% file: Sections/2_Related_Works.tex
\section{Related Works}
\label{sec:related_works}

\subsection{Efficiency Optimization for LLM Agents}
\label{subsec:efficiency_optimization_for_llm_agents}
Large language model (LLM) agents have emerged as a powerful paradigm for complex tasks involving multi-step reasoning, tool invocation, and environment interaction. Recent LLMs with great agentic ability such as Qwen-3~\citep{DBLP:journals/corr/abs-2505-09388} and GPT-OSS~\citep{DBLP:journals/corr/abs-2508-10925}, and representative agentic workflow such as ReAct~\citep{DBLP:conf/iclr/YaoZYDSN023} and Reflexion~\citep{DBLP:conf/nips/ShinnCGNY23} built on top of modern LLM serving engines enable agents to iteratively generate actions and process observations. While these systems demonstrate strong capabilities, their inference efficiency remains a major bottleneck due to iterative generation and repeated model invocation~\citep{DBLP:journals/corr/abs-2508-02694, DBLP:journals/corr/abs-2502-13965}.

Several lines of work aim to improve the efficiency of LLM agent execution.
Some approaches reduce the number of model calls by improving agent-level planning or action selection. For example, LEAP~\citep{verma-bharadwaj-2025-leap} employs look-ahead planning to avoid ineffective actions, while Efficient Agents~\citep{DBLP:journals/corr/abs-2508-02694} optimizes task decomposition and action selection to solve tasks with fewer execution steps.
Others focus on caching or reusing intermediate results across agent steps. For instance, APC~\citep{zhang2025agentic} reuses high-level plans across semantically similar tasks to amortize planning overhead. While these methods significantly improve efficiency, they may also negatively impact generation quality.

\subsection{Speculative Decoding}
\label{subsec:specualtive_decoding}
Speculative decoding is a lossless inference acceleration technique that reduces LLM decoding latency by speculatively generating multiple tokens and verifying them with the target model~\citep{DBLP:conf/acl/XiaYDW00L0S24}.

A common approach, adopted by EAGLE-3~\citep{DBLP:journals/corr/abs-2503-01840} and Multi-Token Prediction (MTP)~\citep{DBLP:journals/corr/abs-2505-07608,DBLP:journals/corr/abs-2412-19437}, trains a lightweight draft model to propose speculative tokens.
More recently, draft-model-free methods retrieve candidate tokens directly from generation history. For example, NGram~\citep{saxena2023prompt} and SuffixDecoding~\citep{oliaro2025suffixdecoding} construct draft continuations by matching n-gram or suffix patterns, eliminating draft inference overhead. However, these algorithm-level designs often suffer from high rejection rates, leading to a high verification cost and poor scalability in large batches. 
Another line of work studies system-level optimizations, such as dynamic batching and budget control~\citep{DBLP:journals/corr/abs-2305-09781,10.1145/3772052.3772239,liu2024optimizing}. For example, 
SPIRe~\citep{DBLP:journals/corr/abs-2504-06419} adapts speculative decoding based on online performance feedback, while AdaSpec~\citep{10.1145/3772052.3772239} models speculative inefficiency to satisfy SLO constraints.
MagicDec~\citep{DBLP:conf/iclr/SadhukhanCCTLSY25} shows that speculative decoding can improve throughput in the long-context scenario. Nevertheless, most of these methods either overlook batch-level acceptance variance or rely on global acceptance statistics, making them brittle to dynamic online workloads.
Consequently, although existing methods achieve notable gains in small-batch settings, their effectiveness under large-batch inference, prevalent in modern LLM serving and agentic applications, remains insufficiently explored.

%% file: Sections/3_Analysis.tex
\section{Speculative Decoding for Batch Inference of LLM Agents}
\label{sec:bottleneck}
In this section, we provide a systematic efficiency analysis of large-batch speculative decoding in the LLM agent scenario. We choose the Code Generation Agent implemented by Reflexion~\citep{DBLP:conf/nips/ShinnCGNY23} and tested on USACO dataset~\citep{DBLP:journals/corr/abs-2404-10952} as the agent workload for analysis. To ensure a fair and realistic analysis, we deploy the LLM of the agent application in vLLM~\citep{DBLP:conf/sosp/KwonLZ0ZY0ZS23}, one of the production-ready LLM serving engines and select two representative speculative decoding algorithms that cover two main types as mentioned in~\cref{subsec:specualtive_decoding}, including the draft-model-based method EAGLE-3~\citep{DBLP:journals/corr/abs-2503-01840} and the draft-model-free method NGram~\citep{saxena2023prompt}. Following the standard of previous work~\citep{DBLP:conf/sosp/KwonLZ0ZY0ZS23}, we choose the total token number generated divided by the total execution time of the workload as the throughput metric for efficiency evaluation.

\subsection{Overall Efficiency Analysis} 
\begin{figure}[t]
\centering
\includegraphics[width=0.48\textwidth]{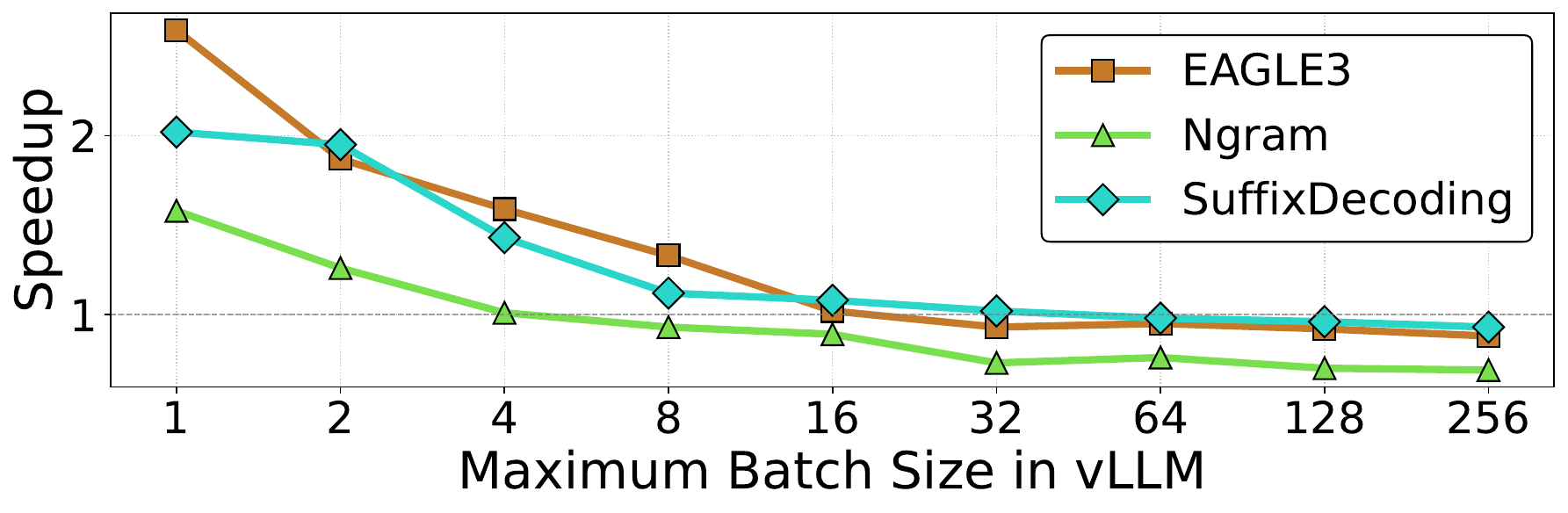}
\caption{Speedup in terms of Goodput (number of generated tokens divided by the execution time)~\citep{liu2024optimizing} under various maximum batch size in vLLM for different speculative decoding algorithms on USACO dataset with Code Generation Agent implemented by Reflexion agentic workflow on Qwen-3-8B. }
\label{fig:moti_bz}
\end{figure}

We first compare the throughput speedup of two representative speculative decoding methods for code generation agents under varying maximum batch sizes in the vLLM engine. The results are shown in~\cref{fig:moti_bz}. When the maximum batch size is 1, EAGLE-3 achieves over $2.5\times$ speedup and SuffixDecoding achieves over $1.9\times$ speedup, consistent with consensus that speculative decoding is effective under small batch sizes. However, as the batch size increases, the speedup rapidly diminishes. When the maximum batch size exceeds 32, speculative decoding even becomes slower than standard autoregressive decoding, indicating that existing speculative decoding methods scale poorly to large-batch agent serving scenarios.

To understand the cause of this degradation, we model the per-step execution time saving $\Delta T(b)$ of speculative decoding with batch size $b$ as:
\begin{align*}
\Delta T(b) &= T_{\text{base}}(b) - T_{\text{spec}}(b) \\
&\approx D(b)(1-\rho(b))t_A - D(b)\big(t_D(b) + t_V(b)\big) \\
&= D(b)\left[(1-\rho(b))t_A - t_D(b) - t_V(b)\right] \\
&\approx D(b)\left[(1-\rho(b))t_A - t_V(b)\right]
\end{align*}
where $D(b)$ denotes the number of drafted tokens in the current step, $\rho(b)$ is the rejection rate that is defined as the ratio of rejected tokens to proposed tokens for each request in the batch, $t_A$ is the per-token time saved by accepting a draft token, and $t_D(b)$ and $t_V(b)$ are the per-token draft and verification costs.

Since $t_D(b)$ is small and can be ignored, this formulation highlights two key factors that impact the efficiency: the rejection rate $\rho(b)$ and $D(b)\big(t_A - t_V(b)\big)$, which we call the utilization of the draft token budget. In general lower rejection rates and higher budget utilization bring in an ideal speculative decoding with larger speedups.

\subsection{Bottleneck Analysis} Motivated by the analysis above, we next examine how existing methods behave with respect to the rejection rate and the token budget utilization.

\noindent\textbf{Bottleneck of High Rejection Rate.} We record the rejection rate, defined as the ratio of rejected tokens to proposed tokens for each request in the batch from vLLM logs during Code Generation Agent inference. ~\cref{tab:rejection_rate} records the average rejection rate for different speculative decoding algorithms. As shown, existing speculative decoding methods suffer from extremely high rejection rates: EAGLE-3 exceeds 50\%, while NGram exceeds 85\% across most batch sizes. Such high rejection rates substantially limit the effective utilization of the draft token budget, directly contributing to the observed degradation in throughput speedup under large-batch settings for LLM agent inference.

\rejTable

\noindent\textbf{Bottleneck of Token Budget Under-Utilization.} We define the available token budget under batch size $b$ as $M(b) = AI - b$, where $AI$ denotes the arithmetic intensity of the deployed hardware for the LLM agent. This definition reflects the fact that, as the batch size increases, LLM inference—particularly the FFN layers—quickly becomes compute-bound and dominates end-to-end latency. As a result, the verification cost $t_V(b)$ increases and eventually approaches the per-token acceptance saving $t_A$, making the total execution time saving $\Delta T(b)$ become:
\begin{align*}
\Delta T(b)
&\approx D(b)\left[(1-\rho(b)) t_A - t_V(b)\right] \\
&= D(b)\left[(1-\rho(b)) t_A - t_A\right] = -D(b)\rho(b) t_A < 0
\end{align*}

\begin{figure}[t]
\centering
\includegraphics[width=0.48\textwidth]{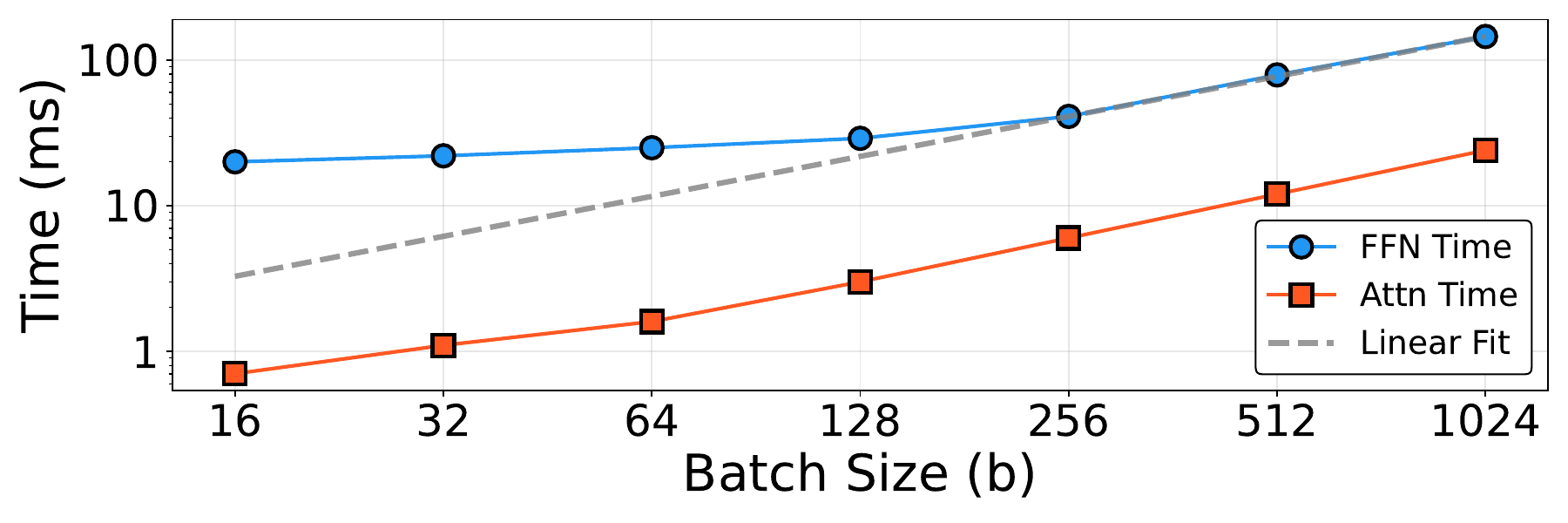}
\caption{Execution time comparison for FFN and Attention modules of Qwen-3-8B under various batch size. }
\label{fig:moti_ffn_attn}
\end{figure}
To further illustrate this effect, we measure the per-layer decoding time of the FFN and attention modules of Qwen-3-8B using vLLM with FlashAttention-2 on an A100 GPU. The results are shown in~\cref{fig:moti_ffn_attn}. When the batch size exceeds 256, which matches the arithmetic intensity of the A100, the FFN latency grows linearly with batch size and quickly dominates the total decoding time. In this regime, decoding becomes compute-bound by the FFN, and the per-token verification cost $t_V(b)$ approaches the per-token autoregressive decoding cost $t_A$. As a result, the time saved by accepting one draft token is almost entirely offset by the cost of verifying it, leaving speculative decoding with little or no performance gain. This observation justifies our definition of the available token budget $M(b)$, which captures the diminishing headroom for speculative tokens as batch size increases.

\begin{figure}[t]
\centering
\includegraphics[width=0.48\textwidth]{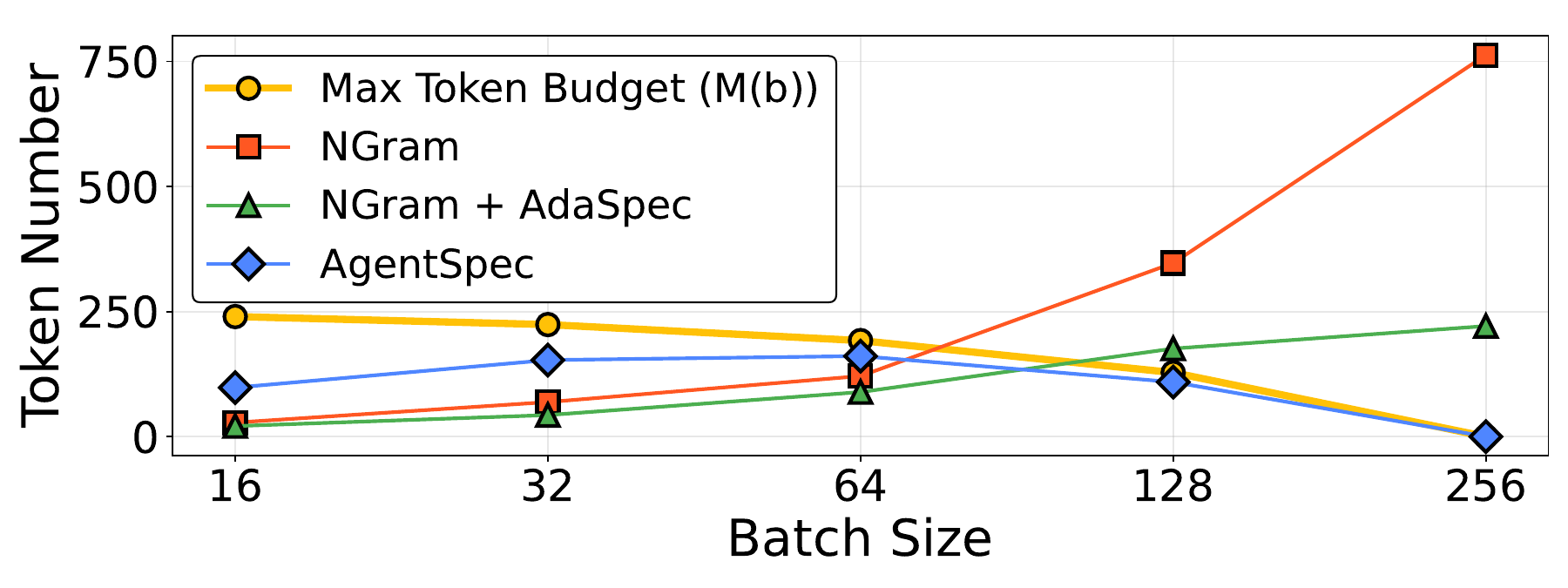}
\caption{Comparison between maximum token budget $M(b)$ and average proposed token number for batch inference of speculative decoding on Reflexion workflow.}
\label{fig:moti_token_number}
\end{figure}
We next record the total number of draft tokens generated per decoding step under different batch sizes and compare it with the available token budget $M(b)$. For a fair comparison, we also implement the budget allocation strategy from AdaSpec~\citep{10.1145/3772052.3772239}, which assigns draft tokens based on per-request acceptance rates. As shown in~\cref{fig:moti_token_number}, we have two key findings. (1) Existing speculative decoding methods severely under-utilize the available token budget, with the actual draft token count remaining far below or over $M(b)$ across all batch sizes.
(2) Existing allocation strategy fails to improve budget utilization, indicating that acceptance-rate–based budgeting is ineffective to improve the batch inference of speculative decoding for LLM agent.

%% file: Sections/4_Method.tex
\section{Method: AgentSpec}
\begin{figure}[ht]
\centering
\includegraphics[width=0.48\textwidth]{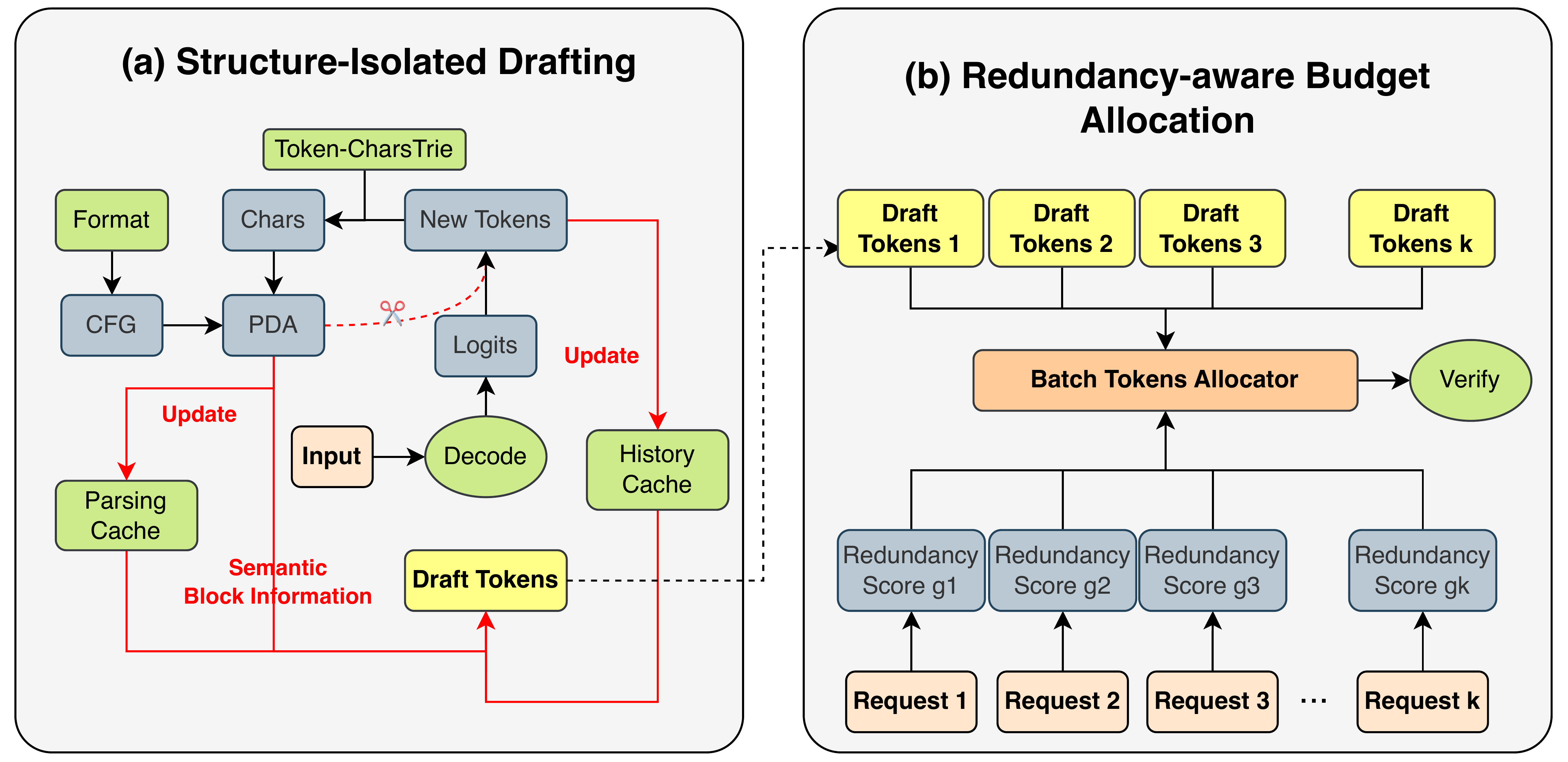}
\caption{Overview of \sysname.}
\label{fig:overview}
\end{figure}
\label{sec:method}
The overview of \sysname is provided in~\cref{fig:overview}. At a high level, \sysname is a model-free speculative decoding algorithm designed for LLM agents. It requires the agent application to provide an agentic structure identifier alongside the input prompt to the LLM server, which enables the system to organize and retrieve historical context according to agent-specific execution structure.
During speculative drafting, \sysname retrieves draft candidates only from historical segments that belong to the same semantic group, thereby avoiding speculation across structurally mismatched contexts. Before verification, \sysname computes an integrated redundancy value for each request in the batch by combining local draft information with global generation history. It then allocates speculative token budgets across requests according to their relative redundancy scores and adjusts the draft length of each request accordingly.
In the following, we describe the two key components of \sysname—structure-isolated drafting and redundancy-aware budget allocation—in detail.

\subsection{Structure-Isolated Drafting}

\label{subsec:structured_isolated_drafting}
\noindent\textbf{Motivation:}
A key distinction between LLM agent applications and standard LLM workloads is that a single user query often spans multiple requests associated with different semantic blocks (e.g., reasoning, tool execution, and result interpretation). While generation patterns are highly consistent within the same semantic block, transitions across blocks or queries introduce substantial semantic shifts.

\begin{figure}[t]
\centering
\includegraphics[width=0.48\textwidth]{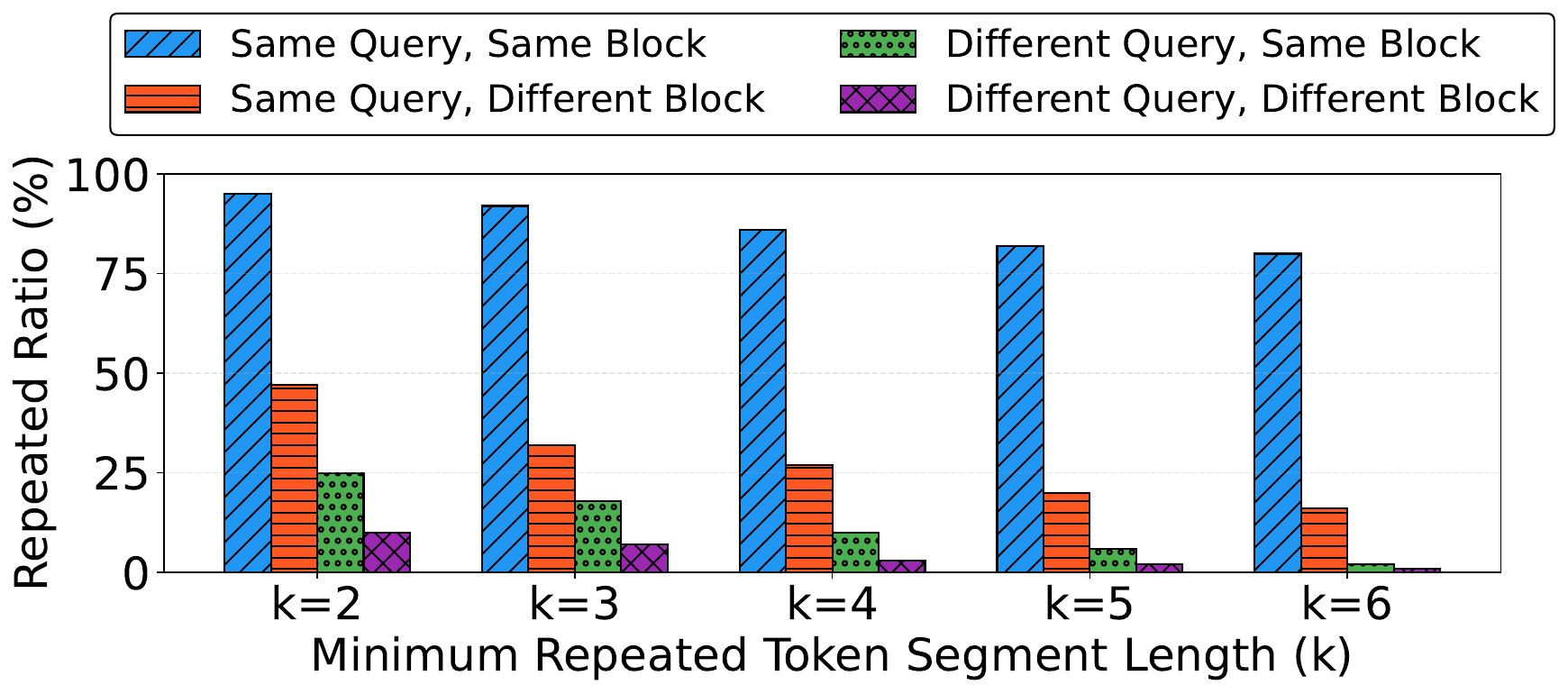}
\caption{Repeated token ratio under different minimum repeated span lengths $k$. The repeated ratio is defined as the total length of maximal repeated segments with at least $k$ consecutive tokens appearing in historical generations, normalized by the total generation length. Repetitions are categorized by whether they occur under the same or different user queries and within the same or different semantic blocks.}

\label{fig:moti_repeated_ratio}
\end{figure}

To quantify this effect, we analyze generation trajectories of Qwen-3-8B on the USACO dataset using a Reflexion-based code generation agent. For each request, we measure repeated token segments that reappear in previously generated content, distinguishing repetitions within the same block from those across different blocks or queries. A repetition is counted only if it contains at least $k$ consecutive tokens, with overlapping matches merged into their longest contiguous span. As shown in~\cref{fig:moti_repeated_ratio}, repeated token segments overwhelmingly occur within the same semantic block of a single query, whereas cross-block and cross-query repetitions are rare. This observation indicates that speculative drafting can be substantially improved by being aware of semantic block boundaries and query scope, thereby reducing unnecessary drafting and speculative rejections. In contrast, existing methods such as NGram and SuffixDecoding retrieve draft candidates from global generation history without such distinctions, leading to high rejection rates and wasted verification cost in agentic workloads.

\noindent\textbf{Key Design:}
The pseudocode of structured-isolated draft of \sysname is provided in~\cref{algo:structure_isolated_drafting}. As shown, the algorithm requires the agent application to explicitly provide semantic structure identifier $S_i$ together with each generation request $r_i$ as below:
$$S_i=\{a_i, q_i, B_i\}$$
where $a_i$ denotes the agent application identifier, $q_i$ denotes the user query index that triggers the current request, and $b_i$ represents the list of $n$ semantic blocks $B_i = [b_i^1, b_i^2, b_i^3,...,b_i^n]$, defined by pairs of start and end string tags identified from previous generation history.

Mapping string-level semantic blocks to token-level generation is challenging due to subword tokenization, where token boundaries do not align with string boundaries. Direct token–string conversion during decoding is prohibitively expensive in common serving engines such as vLLM and sglang, where tokenization and decoding are handled by separate components.

To address this, \sysname adopts a design similar to XGrammar~\citep{DBLP:conf/mlsys/DongRCXZL025} by maintaining a cached token-to-string mapping $M$, enabling efficient online conversion of generated tokens without invoking the tokenizer. Based on this mapping, \sysname maintains a lightweight string-based pushdown automaton (PDA) $P$ on the server, which is incrementally updated to track the current semantic block.

For each request $r_i$, newly generated tokens are converted using $M$ and fed into the corresponding PDA $P(a_i, q_i)$ to identify the active semantic block $b_i^k$. \sysname maintains a separate structure-isolated cache for each semantic block and retrieves speculative drafts exclusively from the matched block, avoiding semantically irrelevant contexts and substantially reducing speculative rejections. As shown in~\cref{tab:rejection_rate}, \sysname achieves a rejection rate as low as 26\%, over 2$\times$ lower than existing baselines.

\subsection{Redundancy-aware Budget Allocation}
\label{subsec:redundancy_aware_budget_allocation}
\begin{figure}[t]
\centering
\includegraphics[width=0.48\textwidth]{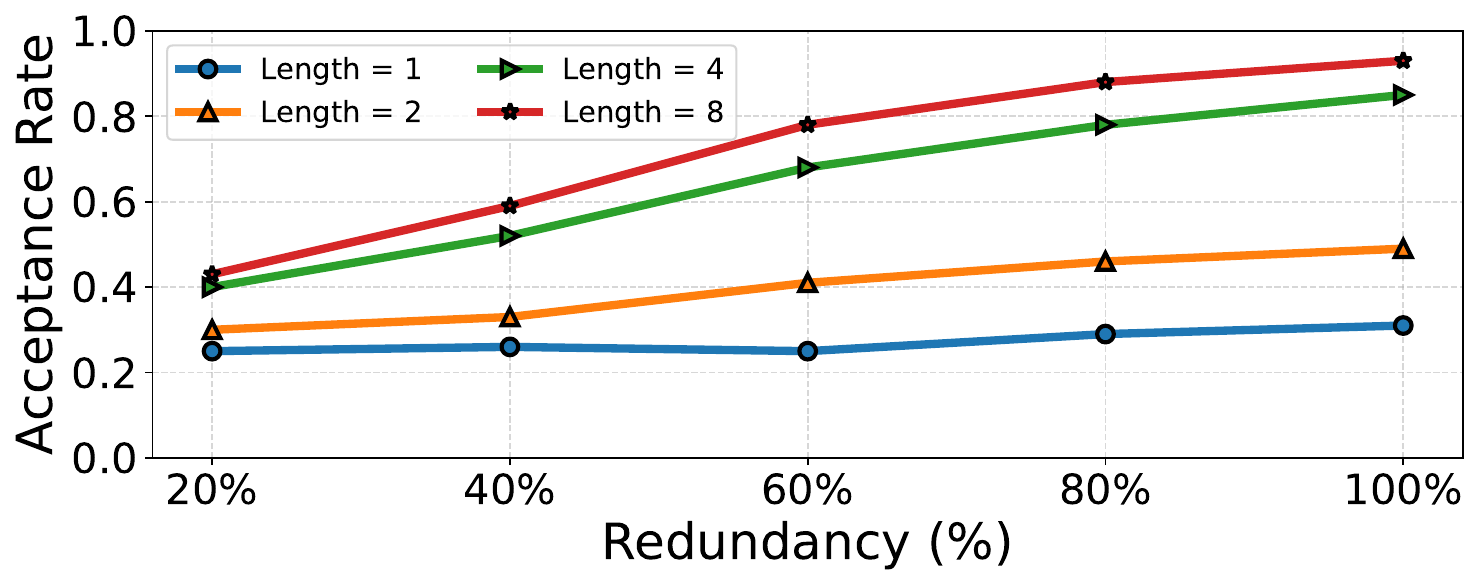}
\caption{Acceptance ratio ($1-\rho(b)$ under different redundancy ratio with different matching pattern length.}
\label{fig:redundancy_accept_map}
\end{figure}

\noindent\textbf{Motivation:} The pseudocode of the redundancy-aware budget allocation is shown in~\cref{algo:redundancy_aware_budget}.
To efficiently utilize the dynamic token budget in speculative decoding, more draft tokens should be assigned to requests with higher acceptance potential. Over-allocating tokens to low-acceptance requests wastes verification cost, while overly conservative allocation under-utilizes the available budget. The key challenge is thus to identify a reliable request-level signal that predicts draft acceptance.
In agent workloads, requests exhibit varying degrees of redundancy in their generation history. Intuitively, drafts aligned with repeated historical patterns are more likely to be accepted. To validate this intuition, we analyze generation traces from Qwen-3-8B on Code Generation Agent and measure draft redundancy as the fraction of matched historical continuations that fully accept the draft. As shown in~\cref{fig:redundancy_accept_map}, higher redundancy strongly correlates with higher acceptance rates, with the correlation strengthening as more matched continuations are observed.

\noindent\textbf{Key Design:} 
To estimate the acceptance likelihood of a speculative draft for each request,
\sysname introduces a redundancy score $g$:
\begin{equation}
g(c,n) = \frac{c}{n}\cdot p(n), \;
p(n) = g_{\min} + (1-g_{\min})(1-e^{1-n}),
\end{equation}
where $n$ is the number of candidate continuations retrieved from the generation
history under the same semantic block and user query, and $c$ is the number of
continuations agreeing on the most frequent prefix. The ratio $c/n$ captures the
consensus among candidates, while the saturation term $p(n)$ downweights unreliable
estimates when historical support is limited. The hyperparameter $g_{\min}$ controls
the minimum confidence under low-support scenarios.

Given the redundancy scores, \sysname computes the batch-level speculative token
budget as
\begin{equation}
B_t = \left\lfloor \frac{\alpha}{bz} \right\rfloor,
\end{equation}
where $bz$ is the batch size and $\alpha$ controls the overall speculative budget.
For each request $r_i$, \sysname retrieves candidate drafts within the same semantic
block and query, identifies the most frequent continuation prefix $CT_i$, and
computes its redundancy score $g_i$. The per-request draft length is then allocated as
\begin{equation}
L_i = \max\!\left(|CT_i|,\; B_t \cdot \frac{g_i}{\sum_j g_j}\right),
\end{equation}
prioritizing requests with higher redundancy while ensuring that each request can
draft at least its most confident prefix.

As shown in~\cref{fig:moti_token_number}, \sysname consistently keeps the total number
of drafted tokens below the maximum budget $M(b)$ and approaches it as batch size
increases, indicating effective utilization of the available speculative budget.

%% file: Sections/5_Experiment.tex
\section{Experiments}
\label{sec:experiments}
\eeTable

\begin{figure*}[t]
\centering
\includegraphics[width=0.98\textwidth]{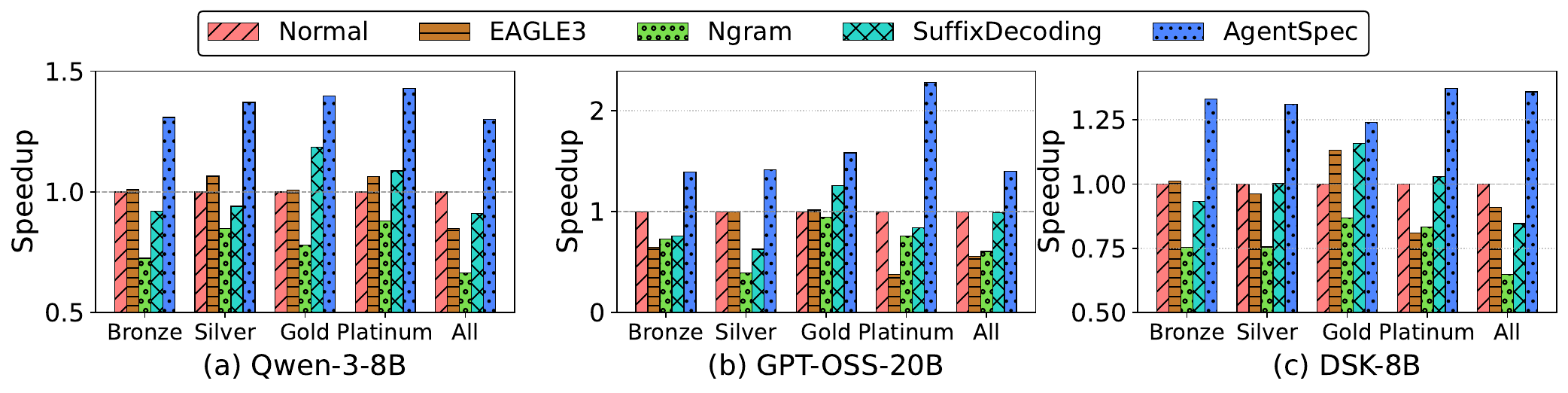}
\vspace{-3mm}
\caption{Speedup in terms of Goodput (token/s) for \sysname and baselines compared with normal autoregressive decoding on all subsets in USACO dataset with Code Generation Agent implemented by Reflexion agentic workflow on three different models.}
\vspace{-3mm}
\label{fig:e2e_subset}
\end{figure*}

\subsection{Experimental Setups}
\label{subsec:setup}

\noindent\textbf{Baselines.} We compare \sysname against two groups of methods: (1) Draft-model-based speculative decoding methods: EAGLE-3~\citep{DBLP:journals/corr/abs-2503-01840} and MTP~\citep{DBLP:journals/corr/abs-2505-07608}. (2) Draft-model-free speculative decoding methods: NGram~\citep{saxena2023prompt} and SuffixDecoding~\citep{oliaro2025suffixdecoding}. All baselines are evaluated under identical serving configurations to ensure a fair comparison.

\noindent\textbf{Models and Datasets.} 
To demonstrate the generality of \sysname, we evaluate its performance on four models from different families, including Qwen-3-8B~\citep{DBLP:journals/corr/abs-2505-09388}, GPT-OSS-20B~\citep{DBLP:journals/corr/abs-2508-10925}, Deepseek-R1-Distill-Llama-8B~\citep{DBLP:journals/corr/abs-2501-12948}, and MiMo-7B~\citep{DBLP:journals/corr/abs-2505-07608}.
We evaluate on four most popular agentic workloads covering two main types. The first is workflow-based agent workload, including Code Generation that is implemented with Reflexion framework~\citep{DBLP:conf/nips/ShinnCGNY23} and tested on USACO benchmark~\citep{DBLP:journals/corr/abs-2404-10952} with multiple difficulty subsets (Bronze, Silver, Gold, and Platinum), and Deep Research that is implemented with LangChain DeepResearch framework~\citep{langchain2026dr} and tested on DeepResearch-Bench benchmark~\citep{DBLP:journals/corr/abs-2506-11763} with three different levels of tasks that require long-horizon reasoning and document synthesis. 
The second is model-based agentic workloads, including SWE-Bench-Lite~\citep{DBLP:conf/iclr/JimenezYWYPPN24} and GAIA~\citep{DBLP:conf/iclr/MialonF0LS24} on OpenHands platform~\citep{DBLP:conf/iclr/0001LSXTZPSLSTL25}.
Finally, we evaluate \sysname on Spec-Bench~\citep{DBLP:conf/acl/XiaYDW00L0S24} to test its performance enhancement on non-agentic data.

\noindent\textbf{Metrics.} Following the standards of Turbospec~\citep{liu2024optimizing}, we use goodput, defined as the total number of generated tokens divided by the execution time of the entire agent workload, as the main metric to evaluate the goodput of \sysname and its baseline methods. We also report speedup in terms of goodput for a clearer comparison, as well as latency~\citep{DBLP:journals/corr/abs-2502-13965}, which is defined as the end-to-end execution time from when a user query enters the agent environment to when the final response is returned.

\noindent\textbf{Implementation Details.} To ensure a fair and realistic comparison, we implemented \sysname and evaluate its performance with baseline methods in vLLM~\citep{DBLP:conf/sosp/KwonLZ0ZY0ZS23}, a production-style LLM serving engine. All other baseline methods are also evaluated with the official implementation provided in vLLM under maximum batch size 256. The detailed configurations in vLLM for running the evaluation is provided in~\cref{appendix:vllm_configration_details}. 

\subsection{End-to-End Comparison}
\label{subsec:end_to_end_comparison}
We first compare the end-to-end efficiency of \sysname against existing baseline methods on two agent applications—Deep Research Agent and Code Generation Agent—as well as two agent benchmarks, GAIA and SWE-Bench, across three different LLMs: Qwen-3-8B, GPT-OSS-20B, and DeepSeek-Distill-LLaMA-8B.
The results are shown in~\cref{tab:e2e_comparison}. \sysname consistently outperforms all baseline speculative decoding methods in terms of efficiency, achieving up to $104\%$ higher goodput. Notably, the goodput of most baseline methods is even lower than that of standard autoregressive decoding, indicating limited benefits in agent workloads.
In contrast, \sysname consistently delivers improved goodput across all agent workloads and model settings, attaining up to a 2.02$\times$ speedup over autoregressive decoding.

\subsection{Comparison with Different Execution Patterns} Even within the same agent workload, problem difficulty can lead to distinct agent execution patterns. A common phenomenon is that more challenging problems induce longer contexts and a larger number of generation steps, as the agent needs to issue more LLM requests to complete the task. To demonstrate the generality of \sysname, we evaluate its efficiency using the Code Generation Agent across all subsets of the USACO dataset, with difficulty levels ranging from Bronze to Platinum. As shown in~\cref{fig:e2e_subset}, \sysname consistently achieves better speedup than other speculative decoding methods across all difficulty subsets and LLMs. More importantly, \sysname also consistently outperforms standard autoregressive decoding, achieving up to a 2.2$\times$ speedup on the USACO subsets.

\subsection{Comparison on Non-Agentic Benchmark}
\label{subsec:spec_bench}
Following the experimental protocol of~\citep{oliaro2025suffixdecoding}, we conduct the evaluation on a non-agentic benchmark Spec-Bench~\citep{DBLP:conf/acl/XiaYDW00L0S24} using Qwen-3-8B. Unlike agentic workloads, which often involve multi-turn requests for a single query and contain explicit semantic structures, Spec-Bench contains fewer repeated historical generations, making it more challenging to utilize history information for drafting. However, as shown in~\cref{tab:qwen_specbench}, \sysname still consistently achieves higher efficiency than both speculative decoding baselines and standard autoregressive decoding across all subsets as well as the full Spec-Bench benchmark. 

\mimoTable

\specbenchTable

\subsection{Comparison with Multi-Token Prediction}
\label{subsec:mtp_comparison}
Multi-Token Prediction (MTP) is a recent speculative decoding paradigm that jointly trains an MTP module as the draft model during the pre-training stage of the target model. To further demonstrate the effectiveness of \sysname, we compare its efficiency with MTP speculative decoding method using MiMo-7B on the Code Generation Agent over the USACO dataset. As shown in~\cref{tab:mimo_usaco}, MTP exhibits limited efficiency gains under batched LLM agent inference. In contrast, \sysname consistently achieves higher efficiency than MTP and maintains a clear speedup over standard autoregressive decoding on MiMo-7B.

\begin{figure}[t]
\centering
\includegraphics[width=0.48\textwidth]{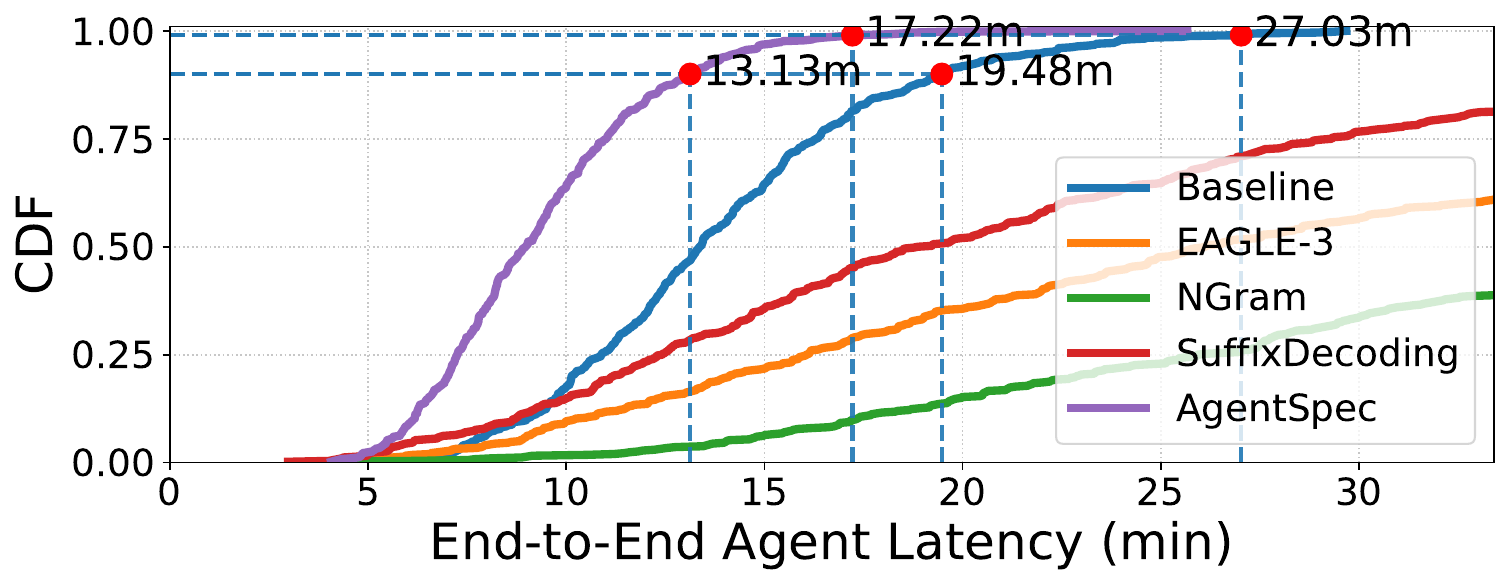}
\caption{Tailed latency analysis for \sysname and baselines on Code Gen agentic workflow.}
\label{fig:e2e_latency}
\end{figure}


\breakdownTable 

\subsection{Latency Analysis}
\label{subsec:tailed_latency_analysis} 
\noindent\textbf{Tailed Latency.} We further analyze the end-to-end program latency, defined as the time elapsed from when a user query is submitted to the agent application to when the final output is returned. The CDF of the tail latency is shown in~\cref{fig:e2e_latency}.
\sysname consistently achieves lower latency than baseline speculative decoding methods, and reduces tail latency by up to 1.47$\times$ and 1.39$\times$ compared to standard autoregressive decoding at the P90 and P99 percentiles, respectively.

\noindent\textbf{Execution Breakdown Analysis.}  We have performed fine-grained analysis including both drafting and verification costs on Reflexion Agentic Workflow. The results in~\cref{tab:runtime_breakdown} show that \sysname significantly reduces verification overhead compared to baselines, which is the dominant factor in overall speedup.  \sysname incurs slightly higher drafting overhead than NGram and SuffixDecoding. This additional cost mainly stems from maintaining the PDA in the structure-isolated drafting component and computing redundancy statistics in the redundancy-aware budget allocation module.
Nevertheless, the overhead remains negligible, amounting to less than 2 ms, and the overall drafting cost of \sysname is still lower than all baseline methods.

\subsection{Ablation Studies}
\label{subsec:ablation_study}
\sensitivityTable

\textbf{Modular Sensitivity Study.} We evaluate the separate contributions of the two key components (i.e., structure-isolated drafting and redundancy-aware budget allocation) of \sysname. 
Let \sysname \texttt{(S)} denote the version of \sysname with structure-isolated drafting only; \sysname \texttt{(R)} denote the version of \sysname with SuffixDecoding and redundancy-aware budget allocation.
As shown in~\cref{tab:sensitivity}, we have three observations. 
(1) \sysname\texttt{(S)}, \sysname\texttt{(R)} and \sysname consistently outperform existing speculative decoding methods. 
(2) \sysname consistently outperforms both \sysname\texttt{(S)} and \sysname\texttt{(R)}.
This result demonstrates the unique contribution from each of the two key components and the importance of combining both components to achieve the best performance.
(3) Comparing between \sysname\texttt{(S)} and \sysname\texttt{(R)}, \sysname\texttt{(S)} achieves a higher speedup compared to \sysname\texttt{(R)}, indicating that structure-isolated drafting plays a more significant role than redundancy-aware budget allocation.

\noindent\textbf{Performance without Explicit Semantic Structures.} Since the structure-isolated drafting component of \sysname requires the agent application to provide explicit semantic block boundaries, some real-world agentic workloads may not expose such structure during generation. To assess the generality of \sysname, we have evaluated a variant that operates without semantic structure inputs, denoted as \sysname\texttt{(N)}. As shown in~\cref{tab:sensitivity}, \sysname\texttt{(N)} consistently outperforms both speculative decoding baselines and standard autoregressive decoding.

\begin{figure}[t]
\centering
\includegraphics[width=\linewidth]{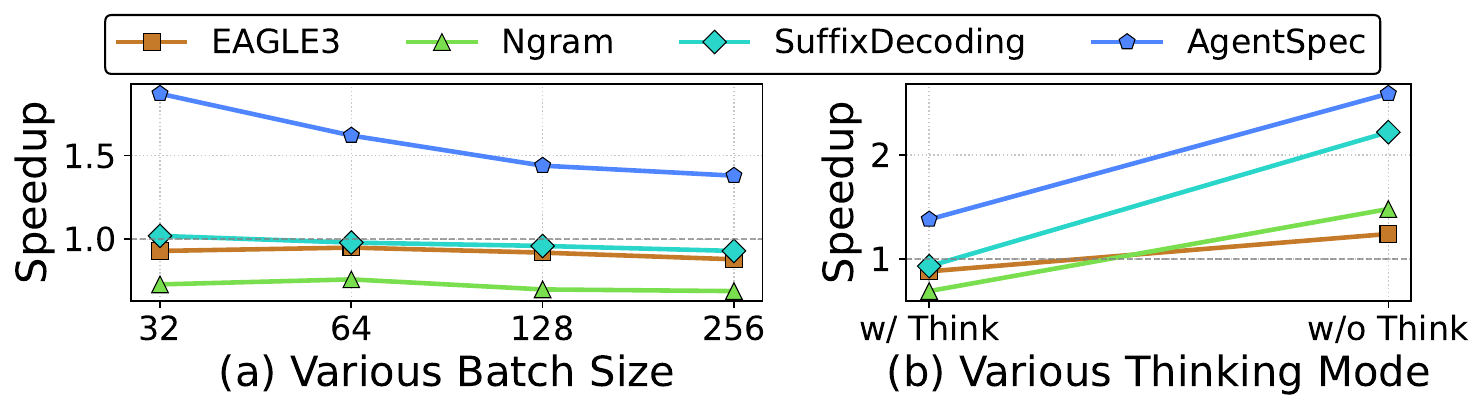}
\caption{Speedup in terms of Goodput (tokens/sec) of \sysname and baselines on workload of code generation agent under various maximum batch size and different thinking modes.}
\label{fig:ablation}
\end{figure}

\noindent\textbf{Performance Under Various Maximum Batch Size.} We next evaluate the performance of \sysname under different maximum batch size configurations in the vLLM engine. As shown in~\cref{fig:ablation}(a), while some existing speculative decoding methods (e.g., SuffixDecoding) achieve higher efficiency than standard autoregressive decoding at small batch sizes, their speedup gradually degrades and can even fall below autoregressive decoding as the maximum batch size increases.
In contrast, \sysname consistently maintains superior efficiency across all batch size configurations, outperforming both speculative decoding baselines and standard autoregressive decoding.

\noindent\textbf{Performance in Different Thinking Modes.} Recent reasoning-oriented LLMs allow users to adjust thinking modes, which can substantially affect both the generated content and output length in agent applications.
To study this effect, we evaluate \sysname on Qwen-3-8B under two thinking modes: w/ think and w/o think~\citep{DBLP:journals/corr/abs-2505-09388}.
As shown in~\cref{fig:ablation}(b), \sysname consistently achieves superior performance across both modes. Notably, under the w/o think setting, \sysname is over 2.5$\times$ faster than standard autoregressive decoding.

%% file: Sections/6_Conclusion.tex
\section{Conclusion}
\label{sec:conclusion}

In this paper, we presented \sysname, a speculative decoding algorithm tailored for batch inference of LLM agents.
\sysname introduces structure-isolated drafting to constrain speculation to semantically coherent segments of the agent workflow, achieving extremely low rejection rates. It also proposes redundancy-aware budget allocation to better exploit dynamically available token budgets using agent-level redundancy.
%
Our experimental results demonstrate the superiority of \sysname over state-of-the-art baselines.
%
%

\section*{Limitation} AgentSpec introduces a lightweight interface between the agent and serving system, which may require minor adaptation in practice. In addition, its structure-isolated drafting component benefits from explicit semantic block information. Although AgentSpec can operate without such metadata, its speedup may depend on the amount of repeated generation patterns available in the workload.

\section*{Ethical Considerations}
AgentSpec is a serving-time acceleration method for LLM-based agents. It does not modify model parameters, training data, or the generation objective, and therefore is not intended to change the output distribution or content policy of the underlying model. Existing safety mechanisms for autoregressive decoding, such as content moderation, tool-use control, and deployment restrictions, should remain applicable when AgentSpec is enabled. Improving inference efficiency may reduce the cost of deploying LLM agents at scale, which can benefit practical applications but may also lower the barrier for misuse, such as automated spam generation or unsafe tool-use workflows. We therefore encourage responsible deployment with appropriate safeguards, including rate limiting, permission control, and monitoring. AgentSpec requires only lightweight semantic structure metadata from the agent application; such metadata should describe workflow structure rather than private or sensitive user information. Our experiments use publicly available benchmarks and do not require collecting additional private user data.

%% file: Sections/Apprendix.tex
\newpage
\appendix
\section{Appendix.}
\subsection{Experiment Configuration Details}
\label{appendix:vllm_configration_details}
We implement \sysname and conduct all experiments using the vLLM v0.12.0 V1 engine. For a fair comparison, we directly run the official implementations of all baseline methods with their default configurations provided by vLLM. We run each experiment on NVIDIA A100 80G GPUs with fix the maximum batch size and serving parameters for all methods. Unless otherwise specified, all goodput and speedup numbers are averaged over five times measurements controlled with various seed numbers and under the default maximum batch size 256 in vLLM.

Specifically, for NGram, we set \lstinline|num-speculative-tokens| to 5 and \lstinline|ngram-prompt-lookup-max| to 4; for SuffixDecoding, we set \lstinline|num-speculative-tokens = 32|.

We use FlashAttention-2 as the attention backend and adopt the default maximum batch size in vLLM (256). For MTP, we directly use the official MTP module provided by MiMo-7B-RL during inference.

For \sysname, the speculation length dynamically adapts to the ongoing batch size and therefore does not require any speculation-length hyperparameters at launch time. Instead, \sysname requires the agent application to provide high-level contextual information, including the agent’s semantic structure and the query index of the current request.

Concretely, we extend the vLLM engine with three additional parameters—\lstinline|structure: string|, \lstinline|query-id: int|, and \lstinline|agent-id: int|—which can be passed by the agent application at inference time. The \lstinline|structure| field encodes the paired delimiters of semantic blocks (e.g., code blocks, tool calls, and mathematical expressions) used by \sysname for redundancy-aware speculation. An example API request is shown in Listing~\ref{lst:redundancy}.

\begin{lstlisting}[language=Python, caption={Example of API request command when using \sysname in vLLM}, label={lst:redundancy}]
resp = await client.chat.completions.create(
        model="Qwen/Qwen3-8B",
        messages=[
            {"role": "system", "content": "Respond in Korean."},
            {"role": "user", "content": f"Say hi. (req {i})"},
        ],
        temperature=0.6,
        extra_body={
        "query_id": 1,
        "agent_id": 1,
        # Code Structure and Math Structure
        "structure": "[
            [("```python", "```"), ("<tool_call>", "<\tool_call>")],
            [("\[", "\]"), ("\(", "\)"), ("$$", "$$"), ("$", "$")]
        ]",
    )
\end{lstlisting}

\begin{table}[ht]
    \centering
    \begin{minipage}{.46\textwidth}
        \GPUTable
    \end{minipage}%
\end{table}

\subsection{Algorithm Pseudocode}
\cref{algo:structure_isolated_drafting} shows the pseudocode of the first component of \sysname, Structure-Isolated Drafting. \cref{algo:redundancy_aware_budget} shows the pseudocode of the second component of \sysname, Redundancy-Aware Budget Allocation.
\begin{algorithm}[ht]
\caption{Structure-Isolated Drafting}
\label{algo:structure_isolated_drafting}
\begin{algorithmic}[1]
    \STATE \textbf{Input:} generation request $r_i = \{a_i, q_i, B_i\}$, history cache $\mathcal{H}$, token-to-string map $M$
    \STATE \textbf{Output:} draft token candidates $CT_i$
    
    \STATE Initialize PDA $P(a_i, q_i)$ for request $r_i$
    \STATE $CT_i \gets \emptyset$
    
    \FOR{each newly generated token $t$ in request $r_i$}
        \STATE Convert $t$ to string using cached map $M$
        \STATE Feed converted string incrementally into $P(a_i, q_i)$
        \STATE Determine current semantic block $b_i^k$ using PDA state
    \ENDFOR
    
    \STATE Retrieve structure-isolated cache $\mathcal{H}(a_i, q_i, b_i^k)$
    \IF{$\mathcal{H}(a_i, q_i, b_i^k)$ is empty}
        \STATE \textbf{return} $\emptyset$
    \ENDIF
    
    \FOR{each historical continuation $h \in \mathcal{H}(a_i, q_i, b_i^k)$}
        \STATE Extract candidate draft continuation $c$ from $h$
        \STATE $CT_i \gets CT_i \cup \{c\}$
    \ENDFOR
    
    \STATE \textbf{return} $CT_i$
\end{algorithmic}
\end{algorithm}

\subsection{Comparison On Various GPU Architectures} We evaluated \sysname on H100 GPU using GPT-OSS-20B. The results in~\cref{tab:hardware_speedup} show that \sysname achieves even higher speedups compared to A100, indicating that our conclusions generalize across hardware generations. From a system perspective, our analysis depends primarily on two factors—rejection rate and token budget utilization—which are also not specific to a particular GPU architecture or serving engine. 

\label{appendix:algorithm_pseudocode}
\begin{algorithm}[ht]
\caption{Redundancy-Aware Budget Allocation}
\label{algo:redundancy_aware_budget}
\begin{algorithmic}[1]
    \STATE \textbf{Input:} batch of requests $\{r_i\}_{i=1}^{bz}$, draft candidate sets $\{CT_i\}$, hyperparameter $\alpha$, $g_{\min}$
    \STATE \textbf{Output:} per-request draft length $\{L_i\}$
    
    \STATE Compute total speculative budget $B_t \gets \left\lfloor \alpha / bz \right\rfloor$
    
    \FOR{each request $r_i$ in batch}
        \STATE $n_i \gets |CT_i|$ \COMMENT{number of candidate continuations}
        \IF{$n_i = 0$}
            \STATE $g_i \gets 0$
            \STATE \textbf{continue}
        \ENDIF
        
        \STATE Identify most frequent continuation prefix $CT_i^{*}$
        \STATE $c_i \gets$ number of continuations agreeing on $CT_i^{*}$
        \STATE $p(n_i) \gets g_{\min} + (1 - g_{\min})(1 - e^{-n_i})$
        \STATE $g_i \gets \frac{c_i}{n_i} \cdot p(n_i)$
    \ENDFOR
    
    \STATE $G \gets \sum_i g_i$
    
    \FOR{each request $r_i$ in batch}
        \STATE $L_i \gets \max\left(|CT_i^{*}|,\; B_t \cdot \frac{g_i}{G}\right)$
    \ENDFOR
    
    \STATE \textbf{return} $\{L_i\}$
\end{algorithmic}
\end{algorithm}

\subsection{Theoretical Complexity analysis} We then provide the time and space complexity analysis of \sysname and other baseline methods as follows. Regarding time complexity, for NGram/SuffixDecoding, drafting consists of constant-time lookup followed by candidate extension, with overall cost $O(K \cdot L)$, where $L$ is the draft length. AgentSpec preserves the same dominant cost. The additional components are lightweight: (a) PDA update: $O(1)$ amortized per token, and (b) semantic filtering + redundancy scoring: $O(K)$ per step. Thus, $T = O(K \cdot L) + O(K)$, which has the same asymptotic complexity as NGram with only a small constant-factor overhead. Regarding space complexity (CPU/DRAM), NGram requires $O(H)$ memory to store history and indices. AgentSpec maintains (a) the same history $O(H)$, (b) a semantic-block index $O(B)$, with $B \ll H$, and (c) one PDA state per request $O(R)$, constant-size each. Thus, $S = O(H + B + R) = O(H)$, matching NGram asymptotically with only lightweight metadata overhead.

%% file: Reference.bib
@article{DBLP:journals/tmlr/Wan0LA0LQYZZC024,
  author       = {Zhongwei Wan and
                  Xin Wang and
                  Che Liu and
                  Samiul Alam and
                  Yu Zheng and
                  Jiachen Liu and
                  Zhongnan Qu and
                  Shen Yan and
                  Yi Zhu and
                  Quanlu Zhang and
                  Mosharaf Chowdhury and
                  Mi Zhang},
  title        = {Efficient Large Language Models: {A} Survey},
  journal      = {Trans. Mach. Learn. Res.},
  volume       = {2024},
  year         = {2024}
}

@article{wang2024iot,
    title={IoT in the Era of Generative AI: Vision and Challenges},
    author={Wang, Xin and Wan, Zhongwei and Hekmati, Arvin and Zong, Mingyu and Alam, Samiul and Zhang, Mi and Krishnamachari, Bhaskar},
    journal={arXiv preprint arXiv:2401.01923},
    year={2024}
}

@inproceedings{DBLP:conf/iclr/YaoZYDSN023,
  author       = {Shunyu Yao and
                  Jeffrey Zhao and
                  Dian Yu and
                  Nan Du and
                  Izhak Shafran and
                  Karthik R. Narasimhan and
                  Yuan Cao},
  title        = {ReAct: Synergizing Reasoning and Acting in Language Models},
  booktitle    = {{ICLR}},
  publisher    = {OpenReview.net},
  year         = {2023}
}

@inproceedings{verma-bharadwaj-2025-leap,
    title = "{LEAP} {\&} {LEAN}: Look-ahead Planning and Agile Navigation for {LLM} Agents",
    author = "Verma, Nikhil  and
      Bharadwaj, Manasa",
    editor = "Rehm, Georg  and
      Li, Yunyao",
    booktitle = "Proceedings of the 63rd Annual Meeting of the Association for Computational Linguistics (Volume 6: Industry Track)",
    month = jul,
    year = "2025",
    address = "Vienna, Austria",
    publisher = "Association for Computational Linguistics",
    url = "https://aclanthology.org/2025.acl-industry.64/",
    doi = "10.18653/v1/2025.acl-industry.64",
    pages = "896--933",
    ISBN = "979-8-89176-288-6",
}

@inproceedings{
    zhang2025agentic,
    title={Agentic Plan Caching: Test-Time Memory for Fast and Cost-Efficient {LLM} Agents},
    author={Qizheng Zhang and Michael Wornow and Kunle Olukotun},
    booktitle={The Thirty-ninth Annual Conference on Neural Information Processing Systems},
    year={2025},
    url={https://openreview.net/forum?id=n4V3MSqK77}
}

@article{liu2024optimizing,
  title={Optimizing speculative decoding for serving large language models using goodput},
  author={Liu, Xiaoxuan and Daniel, Cade and Hu, Langxiang and Kwon, Woosuk and Li, Zhuohan and Mo, Xiangxi and Cheung, Alvin and Deng, Zhijie and Stoica, Ion and Zhang, Hao},
  journal={arXiv preprint arXiv:2406.14066},
  year={2024}
}

@article{DBLP:journals/corr/abs-2503-21460,
  author       = {Junyu Luo and
                  Weizhi Zhang and
                  Ye Yuan and
                  Yusheng Zhao and
                  Junwei Yang and
                  Yiyang Gu and
                  Bohan Wu and
                  Binqi Chen and
                  Ziyue Qiao and
                  Qingqing Long and
                  Rongcheng Tu and
                  Xiao Luo and
                  Wei Ju and
                  Zhiping Xiao and
                  Yifan Wang and
                  Meng Xiao and
                  Chenwu Liu and
                  Jingyang Yuan and
                  Shichang Zhang and
                  Yiqiao Jin and
                  Fan Zhang and
                  Xian Wu and
                  Hanqing Zhao and
                  Dacheng Tao and
                  Philip S. Yu and
                  Ming Zhang},
  title        = {Large Language Model Agent: {A} Survey on Methodology, Applications
                  and Challenges},
  journal      = {CoRR},
  volume       = {abs/2503.21460},
  year         = {2025}
}

@inproceedings{DBLP:conf/coling/Li25,
  author       = {Xinzhe Li},
  title        = {A Review of Prominent Paradigms for LLM-Based Agents: Tool Use, Planning
                  (Including RAG), and Feedback Learning},
  booktitle    = {{COLING}},
  pages        = {9760--9779},
  publisher    = {Association for Computational Linguistics},
  year         = {2025}
}

@article{DBLP:journals/corr/abs-2412-19437,
  author       = {DeepSeek{-}AI},
  title        = {DeepSeek-V3 Technical Report},
  journal      = {CoRR},
  volume       = {abs/2412.19437},
  year         = {2024}
}

@inproceedings{DBLP:conf/nips/ZhengYXS0YCKSGB24,
  author       = {Lianmin Zheng and
                  Liangsheng Yin and
                  Zhiqiang Xie and
                  Chuyue Sun and
                  Jeff Huang and
                  Cody Hao Yu and
                  Shiyi Cao and
                  Christos Kozyrakis and
                  Ion Stoica and
                  Joseph E. Gonzalez and
                  Clark W. Barrett and
                  Ying Sheng},
  title        = {SGLang: Efficient Execution of Structured Language Model Programs},
  booktitle    = {NeurIPS},
  year         = {2024}
}

@inproceedings{DBLP:conf/iclr/SadhukhanCCTLSY25,
  author       = {Ranajoy Sadhukhan and
                  Jian Chen and
                  Zhuoming Chen and
                  Vashisth Tiwari and
                  Ruihang Lai and
                  Jinyuan Shi and
                  Ian En{-}Hsu Yen and
                  Avner May and
                  Tianqi Chen and
                  Beidi Chen},
  title        = {MagicDec: Breaking the Latency-Throughput Tradeoff for Long Context
                  Generation with Speculative Decoding},
  booktitle    = {{ICLR}},
  publisher    = {OpenReview.net},
  year         = {2025}
}

@inproceedings{DBLP:conf/acl/WuZVPRL25,
  author       = {Zhaoxuan Wu and
                  Zijian Zhou and
                  Arun Verma and
                  Alok Prakash and
                  Daniela Rus and
                  Bryan Kian Hsiang Low},
  title        = {{TETRIS:} Optimal Draft Token Selection for Batch Speculative Decoding},
  booktitle    = {{ACL} {(1)}},
  pages        = {33329--33345},
  publisher    = {Association for Computational Linguistics},
  year         = {2025}
}

@article{DBLP:journals/corr/abs-2504-06419,
  author       = {Sanjit Neelam and
                  Daniel Heinlein and
                  Vaclav Cvicek and
                  Akshay Mishra and
                  Reiner Pope},
  title        = {SPIRe: Boosting {LLM} Inference Throughput with Speculative Decoding},
  journal      = {CoRR},
  volume       = {abs/2504.06419},
  year         = {2025}
}

@inproceedings{10.1145/3772052.3772239,
    author = {Huang, Kaiyu and Wu, Hao and Shi, Zhubo and Zou, Han and Yu, Minchen and Shi, Qingjiang},
    title = {AdaSpec: Adaptive Speculative Decoding for Fast, SLO-Aware Large Language Model Serving},
    year = {2026},
    isbn = {9798400722769},
    publisher = {Association for Computing Machinery},
    address = {New York, NY, USA},
    url = {https://doi.org/10.1145/3772052.3772239},
    doi = {10.1145/3772052.3772239},
    booktitle = {Proceedings of the 2025 ACM Symposium on Cloud Computing},
    pages = {361–374},
    numpages = {14},
    series = {SoCC '25}
}

@article{DBLP:journals/corr/abs-2305-09781,
  author       = {Xupeng Miao and
                  Gabriele Oliaro and
                  Zhihao Zhang and
                  Xinhao Cheng and
                  Zeyu Wang and
                  Rae Ying Yee Wong and
                  Zhuoming Chen and
                  Daiyaan Arfeen and
                  Reyna Abhyankar and
                  Zhihao Jia},
  title        = {SpecInfer: Accelerating Generative {LLM} Serving with Speculative
                  Inference and Token Tree Verification},
  journal      = {CoRR},
  volume       = {abs/2305.09781},
  year         = {2023}
}

@article{DBLP:journals/corr/abs-2502-13965,
  author       = {Michael Luo and
                  Xiaoxiang Shi and
                  Colin Cai and
                  Tianjun Zhang and
                  Justin Wong and
                  Yichuan Wang and
                  Chi Wang and
                  Yanping Huang and
                  Zhifeng Chen and
                  Joseph E. Gonzalez and
                  Ion Stoica},
  title        = {Autellix: An Efficient Serving Engine for {LLM} Agents as General
                  Programs},
  journal      = {CoRR},
  volume       = {abs/2502.13965},
  year         = {2025}
}

@article{DBLP:journals/corr/abs-2508-02694,
  author       = {Ningning Wang and
                  Xavier Hu and
                  Pai Liu and
                  He Zhu and
                  Yue Hou and
                  Heyuan Huang and
                  Shengyu Zhang and
                  Jian Yang and
                  Jiaheng Liu and
                  Ge Zhang and
                  Changwang Zhang and
                  Jun Wang and
                  Yuchen Eleanor Jiang and
                  Wangchunshu Zhou},
  title        = {Efficient Agents: Building Effective Agents While Reducing Cost},
  journal      = {CoRR},
  volume       = {abs/2508.02694},
  year         = {2025}
}

@misc{saxena2023prompt,
    title = {Prompt Lookup Decoding},
    author = {Apoorv Saxena},
    year = {2023},
    month = {November},
    url = {https://github.com/apoorvumang/prompt-lookup-decoding/}
}

@inproceedings{oliaro2025suffixdecoding,
  author    = {Gabriele Oliaro and Zhihao Jia and Daniel Campos and Aurick Qiao},
  title     = {SuffixDecoding: Extreme Speculative Decoding for Emerging AI Applications},
  booktitle = {The Thirty-ninth Annual Conference on Neural Information Processing Systems},
  year      = {2025},
  url       = {https://arxiv.org/abs/2411.04975}
}

@article{DBLP:journals/corr/abs-2503-01840,
  author       = {Yuhui Li and
                  Fangyun Wei and
                  Chao Zhang and
                  Hongyang Zhang},
  title        = {{EAGLE-3:} Scaling up Inference Acceleration of Large Language Models
                  via Training-Time Test},
  journal      = {CoRR},
  volume       = {abs/2503.01840},
  year         = {2025}
}

@inproceedings{DBLP:conf/mlsys/DongRCXZL025,
  author       = {Yixin Dong and
                  Charlie F. Ruan and
                  Yaxing Cai and
                  Ziyi Xu and
                  Yilong Zhao and
                  Ruihang Lai and
                  Tianqi Chen},
  title        = {XGrammar: Flexible and Efficient Structured Generation Engine for
                  Large Language Models},
  booktitle    = {MLSys},
  publisher    = {OpenReview.net/mlsys.org},
  year         = {2025}
}

@inproceedings{DBLP:conf/sosp/KwonLZ0ZY0ZS23,
  author       = {Woosuk Kwon and
                  Zhuohan Li and
                  Siyuan Zhuang and
                  Ying Sheng and
                  Lianmin Zheng and
                  Cody Hao Yu and
                  Joseph Gonzalez and
                  Hao Zhang and
                  Ion Stoica},
  title        = {Efficient Memory Management for Large Language Model Serving with
                  PagedAttention},
  booktitle    = {{SOSP}},
  pages        = {611--626},
  publisher    = {{ACM}},
  year         = {2023}
}

@inproceedings{DBLP:conf/nips/ShinnCGNY23,
  author       = {Noah Shinn and
                  Federico Cassano and
                  Ashwin Gopinath and
                  Karthik Narasimhan and
                  Shunyu Yao},
  title        = {Reflexion: language agents with verbal reinforcement learning},
  booktitle    = {NeurIPS},
  year         = {2023}
}

@inproceedings{DBLP:conf/iclr/JimenezYWYPPN24,
  author       = {Carlos E. Jimenez and
                  John Yang and
                  Alexander Wettig and
                  Shunyu Yao and
                  Kexin Pei and
                  Ofir Press and
                  Karthik R. Narasimhan},
  title        = {SWE-bench: Can Language Models Resolve Real-world Github Issues?},
  booktitle    = {{ICLR}},
  publisher    = {OpenReview.net},
  year         = {2024}
}

@inproceedings{DBLP:conf/iclr/MialonF0LS24,
  author       = {Gr{\'{e}}goire Mialon and
                  Cl{\'{e}}mentine Fourrier and
                  Thomas Wolf and
                  Yann LeCun and
                  Thomas Scialom},
  title        = {{GAIA:} a benchmark for General {AI} Assistants},
  booktitle    = {{ICLR}},
  publisher    = {OpenReview.net},
  year         = {2024}
}

@inproceedings{DBLP:conf/acl/XiaYDW00L0S24,
  author       = {Heming Xia and
                  Zhe Yang and
                  Qingxiu Dong and
                  Peiyi Wang and
                  Yongqi Li and
                  Tao Ge and
                  Tianyu Liu and
                  Wenjie Li and
                  Zhifang Sui},
  title        = {Unlocking Efficiency in Large Language Model Inference: {A} Comprehensive
                  Survey of Speculative Decoding},
  booktitle    = {{ACL} (Findings)},
  pages        = {7655--7671},
  publisher    = {Association for Computational Linguistics},
  year         = {2024}
}

@article{DBLP:journals/corr/abs-2404-10952,
  author       = {Quan Shi and
                  Michael Tang and
                  Karthik Narasimhan and
                  Shunyu Yao},
  title        = {Can Language Models Solve Olympiad Programming?},
  journal      = {CoRR},
  volume       = {abs/2404.10952},
  year         = {2024}
}

@article{DBLP:journals/corr/abs-2506-11763,
  author       = {Mingxuan Du and
                  Benfeng Xu and
                  Chiwei Zhu and
                  Xiaorui Wang and
                  Zhendong Mao},
  title        = {DeepResearch Bench: {A} Comprehensive Benchmark for Deep Research
                  Agents},
  journal      = {CoRR},
  volume       = {abs/2506.11763},
  year         = {2025}
}

@misc{langchain2026dr,
    title = {Langchain Open Deep Research},
    author = {Langchain Team},
    year = {2026},
    url = {https://github.com/langchain-ai/open_deep_research/}
}

@inproceedings{DBLP:conf/iclr/0001LSXTZPSLSTL25,
  author       = {Xingyao Wang and
                  Boxuan Li and
                  Yufan Song and
                  Frank F. Xu and
                  Xiangru Tang and
                  Mingchen Zhuge and
                  Jiayi Pan and
                  Yueqi Song and
                  Bowen Li and
                  Jaskirat Singh and
                  Hoang H. Tran and
                  Fuqiang Li and
                  Ren Ma and
                  Mingzhang Zheng and
                  Bill Qian and
                  Yanjun Shao and
                  Niklas Muennighoff and
                  Yizhe Zhang and
                  Binyuan Hui and
                  Junyang Lin and
                  et al.},
  title        = {OpenHands: An Open Platform for {AI} Software Developers as Generalist
                  Agents},
  booktitle    = {{ICLR}},
  publisher    = {OpenReview.net},
  year         = {2025}
}

@article{DBLP:journals/corr/abs-2505-09388,
  author       = {An Yang and
                  Anfeng Li and
                  Baosong Yang and
                  Beichen Zhang and
                  Binyuan Hui and
                  Bo Zheng and
                  Bowen Yu and
                  Chang Gao and
                  Chengen Huang and
                  Chenxu Lv and
                  Chujie Zheng and
                  Dayiheng Liu and
                  Fan Zhou and
                  Fei Huang and
                  Feng Hu and
                  Hao Ge and
                  Haoran Wei and
                  Huan Lin and
                  Jialong Tang and
                  Jian Yang and
                  Jianhong Tu and
                  Jianwei Zhang and
                  Jian Yang and
                  Jiaxi Yang and
                  Jingren Zhou and
                  Junyang Lin and
                  Kai Dang and
                  Keqin Bao and
                  Kexin Yang and
                  Le Yu and
                  Lianghao Deng and
                  Mei Li and
                  Mingfeng Xue and
                  Mingze Li and
                  Pei Zhang and
                  Peng Wang and
                  Qin Zhu and
                  Rui Men and
                  Ruize Gao and
                  Shixuan Liu and
                  Shuang Luo and
                  Tianhao Li and
                  Tianyi Tang and
                  Wenbiao Yin and
                  Xingzhang Ren and
                  Xinyu Wang and
                  Xinyu Zhang and
                  Xuancheng Ren and
                  Yang Fan and
                  Yang Su and
                  Yichang Zhang and
                  Yinger Zhang and
                  Yu Wan and
                  Yuqiong Liu and
                  Zekun Wang and
                  Zeyu Cui and
                  Zhenru Zhang and
                  Zhipeng Zhou and
                  Zihan Qiu},
  title        = {Qwen3 Technical Report},
  journal      = {CoRR},
  volume       = {abs/2505.09388},
  year         = {2025}
}

@article{DBLP:journals/corr/abs-2508-10925,
  author       = {OpenAI},
  title        = {gpt-oss-120b {\&} gpt-oss-20b Model Card},
  journal      = {CoRR},
  volume       = {abs/2508.10925},
  year         = {2025}
}

@article{DBLP:journals/corr/abs-2501-12948,
  author       = {DeepSeek{-}AI},
  title        = {DeepSeek-R1: Incentivizing Reasoning Capability in LLMs via Reinforcement
                  Learning},
  journal      = {CoRR},
  volume       = {abs/2501.12948},
  year         = {2025}
}

@article{DBLP:journals/corr/abs-2505-07608,
  author       = {Bingquan Xia and
                  Bowen Shen and
                  Cici and
                  Dawei Zhu and
                  Di Zhang and
                  Gang Wang and
                  Hailin Zhang and
                  Huaqiu Liu and
                  Jiebao Xiao and
                  Jinhao Dong and
                  Liang Zhao and
                  Peidian Li and
                  Peng Wang and
                  Shihua Yu and
                  Shimao Chen and
                  Weikun Wang and
                  Wenhan Ma and
                  Xiangwei Deng and
                  Yi Huang and
                  Yifan Song and
                  Zihan Jiang and
                  Bowen Ye and
                  Can Cai and
                  Chenhong He and
                  Dong Zhang and
                  Duo Zhang and
                  Guoan Wang and
                  Hao Tian and
                  Haochen Zhao and
                  Heng Qu and
                  Hongshen Xu and
                  Jun Shi and
                  Kainan Bao and
                  QingKai Fang and
                  Kang Zhou and
                  Kangyang Zhou and
                  Lei Li and
                  Menghang Zhu and
                  Nuo Chen and
                  Qiantong Wang and
                  Shaohui Liu and
                  Shicheng Li and
                  Shuhao Gu and
                  Shuhuai Ren and
                  Shuo Liu and
                  Sirui Deng and
                  Weiji Zhuang and
                  Weiwei Lv and
                  Wenyu Yang and
                  Xin Zhang and
                  Xing Yong and
                  Xing Zhang and
                  Xingchen Song and
                  Xinzhe Xu and
                  Xu Wang and
                  Yihan Yan and
                  Yu Tu and
                  Yuanyuan Tian and
                  Yudong Wang and
                  Yue Yu and
                  Zhenru Lin and
                  Zhichao Song and
                  Zihao Yue},
  title        = {MiMo: Unlocking the Reasoning Potential of Language Model - From Pretraining
                  to Posttraining},
  journal      = {CoRR},
  volume       = {abs/2505.07608},
  year         = {2025}
}
